\documentclass[journal]{IEEEtran}
\ifCLASSINFOpdf
  \usepackage[pdftex]{graphicx}
  \usepackage{subcaption}
  \usepackage[font={footnotesize}]{caption}
  \usepackage{url}
  \usepackage{xcolor}
  \usepackage{booktabs}
\else
\fi
\usepackage{amsmath}

\begin{document}
\title{Instruction-Driven 3D Facial Expression Generation and Transition}

%%%%%%% NOTE %%%%%%%%
% \author{Michael~Shell,~\IEEEmembership{Member,~IEEE,}
%         John~Doe,~\IEEEmembership{Fellow,~OSA,}
%         and~Jane~Doe,~\IEEEmembership{Life~Fellow,~IEEE}% <-this % stops a space
\author{Anh H. Vo,~\IEEEmembership{}
        Tae-Seok Kim,~\IEEEmembership{}
        Hulin Jin,~\IEEEmembership{}
        Soo-Mi Choi,~\IEEEmembership{}
        and Yong-Guk Kim*~\IEEEmembership{}% <-this % stops a space

\thanks{A H Vo, T-S Kim, S-M Choi, and Y-G Kim are with the Department of Computer Engineering, Sejong University, Seoul, Republic of Korea.}
\thanks{H Jin is with the School of Computer Science and Technology, Anhui University, Hefei, China.}

\thanks{* Corresponding Author: ykim@sejong.ac.kr.}
\thanks{
This is the author’s accepted manuscript.
The final version is published in IEEE Transactions on Multimedia, 2025
}
} % <-this % stops a space


\markboth{Journal of \LaTeX\ Class }%
{Shell \MakeLowercase{\textit{et al.}}: Bare Demo of IEEEtran.cls for IEEE Journals}



\maketitle
%% 0. ABSTRACT %%
\begin{abstract}
 A 3D avatar typically has one of six cardinal facial expressions. To simulate realistic emotional variation, we should be able to render a facial transition between two arbitrary expressions. This study presents a new framework for instruction-driven facial expression generation that produces a 3D face and, starting from an image of the face,  transforms the facial expression from one designated facial expression to another. The Instruction-driven Facial Expression Decomposer (IFED) module is introduced to facilitate multimodal data learning and capture the correlation between textual descriptions and facial expression features. Subsequently, we propose the Instruction to Facial Expression Transition (I2FET) method, which leverages IFED and a vertex reconstruction loss function to refine the semantic comprehension of latent vectors, thus generating a facial expression sequence according to the given instruction. Lastly, we present the Facial Expression Transition model to generate smooth transitions between facial expressions.
Extensive evaluation suggests that the proposed model outperforms state-of-the-art methods on the CK+ and CelebV-HQ datasets. The results show that our framework can generate facial expression trajectories according to text instruction. Considering that text prompts allow us to make diverse descriptions of human emotional states, the repertoire of facial expressions and the transitions between them can be expanded greatly. We expect our framework to find various practical applications. More information about our project can be found at \url{https://vohoanganh.github.io/tg3dfet/}
\end{abstract} 


% \fi

\IEEEpeerreviewmaketitle
   
\section{Introduction}
\label{sec:intro}
\IEEEPARstart{F}ace and facial expression generation have attracted much attention in recent years due to their widespread application in many fields such as augmented reality/virtual reality, the gaming industry, movie production, or human-computer interfaces. Face generation can be divided into two categories depending on whether the face is two or three-dimensional (2D or 3D faces). For the 2D case, several methods \cite{huang2023collaborative, xia2021tedigan, jiang2021talkedit, Kim_2023_CVPR, Susitong2023} have been proposed to generate facial images using multimodal input, such as text, sketches, and segmentation. Previous studies \cite{ding2017exprgan, xia2021tedigan, Guo_faceclip_2023, hang2023lang} have attempted to solve facial expression generation in response to input provided as an emotion label or text prompts. However, these methods have limited ability to control the geometrical information of the output image and are effective only for specific applications. In the case of a 3D face, several studies \cite{Khakhulin2022ROME, INSTA, ma2023cvthead, pengbo2022, chen2023_kbs} have demonstrated methods for generating 3D faces from 2D ones. The models generate the output image combining geometrical information from the source image,  such as the pose, shape, and texture.

\begin{figure}[ht!]
   \belowcaptionskip = -10pt
    \centering
    \includegraphics[scale=0.44]{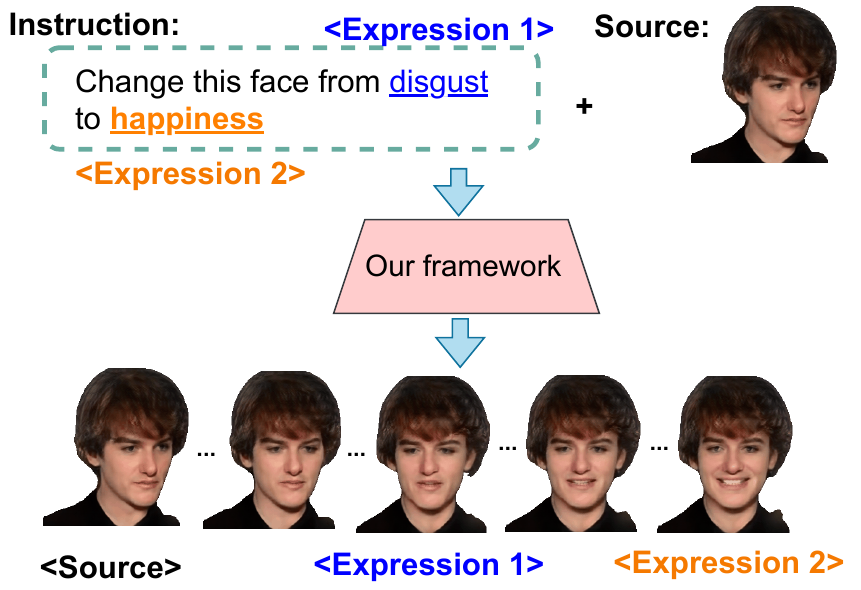}
    \caption{Illustration of our framework wherein it accepts a text instruction with a face image (source) as input and generates a 3D face from it. Then, it transforms the facial expression, from disgust (expression 1) to happiness (expression 2) via a neutral expression.} 
    \label{fig:idea}
\end{figure}

Other researchers \cite{Hwang_2023_ICCV, aneja2023clipface} generate a specific 3D facial expression based on text prompts. They utilize CLIP (Contrastive Language-Image Pretraining) or NeRF (Neural Radiance Field) to learn the latent code for manipulating facial expressions. Yet, NeRF requires substantial computational resources and time. For instance, \cite{aneja2023clipface} learns facial parameters and generates facial texture and expression based on text prompts. Although these methods can generate specific facial expressions, they do not address facial expression transitions based on instruction provided. Meanwhile, other studies \cite{Doukas_2021_ICCV, buehler2021varitex, grassal2022neural, paraperas2022ned, qian2023gaussianavatars} proposed to control facial expressions from an RGB video or generate portrait images \cite{yue2023aniportraitgan} with controllable facial expression, head pose, and shoulder movements. Alternatively, \cite{Otberdout2022GeneratingC4, zou20234d} utilized the facial landmark trajectories to generate multiple 4D expression transitions on a meshed face. 

These studies led us to propose a new framework that enables a 3D face to be generated from a source image. Then, the facial expression undergoes transition in response to prompt text provided by the user to request a facial transition between two arbitrary expressions. Recent studies have used text-driven methods to generate a deformed face via NeRF or create an animated face via StyleGAN \cite{hang2023lang, Hwang_2023_ICCV}. They generate sequence of facial expressions by specifying a certain emotion, such as 'the person is happy'. In contrast, our study utilizes a text instruction to control the variation in the facial expression, such as 'Turn this face from disgust to happiness', as illustrated in Fig. \ref{fig:idea}. Yet, understanding the target expressions specified by text instructions presents a challenge when relying solely on global text embedding. Indeed, even though CLIP is a notable method for the language-conditioning module in the text-to-image task \cite{hang2023lang, Hwang_2023_ICCV, gu2023seer, aneja2023clipface}, it still faces challenges related to bias in understanding the text despite its large-scale training or the provision of global instruction, which may overlook fine-grained sub-instructions. Moreover, a text-only approach could be sensitive in generative models because changes in the structure of instructions can significantly impact their performance.
The additional challenge lies therein to ensure that the facial motion described in the instruction is aligned with the semantics of the language. A straightforward approach to generating the facial motion would entail interpolating between the source code and the edited code from StyleClip \cite{Patashnik_2021_ICCV}. 
Nevertheless, the additional challenge of controlling the latent code to adjust the head pose of the source image while generating the facial animation makes it difficult to achieve the intended goal.
\\
\hspace*{0.3cm}To make the connection between the text and facial expressions, we supplement a text instruction for making a transition between facial expressions into two standard datasets containing facial images, i.e., CelebV-HQ (High-Quality Celebrity Video) and CK+ (Extended Cohn-Kanade). To control the facial model in detail, FLAME (Faces Learned with an Articulated Model and Expressions) \cite{Li2017LearningAM} is adopted and then fine-tuned with basic facial expressions and corresponding poses to capture the movement of the head, eyes, and mouth.\\
\hspace*{0.3cm}Furthermore, integrating feature representations from different modalities minimizes ambiguity in redundant resources and effectively handles missing modalities, particularly in generative models, by improving inter-modality relations \cite{Sutter2020SupplementaryMM, Suzuki2022ASO}. 
The proposed approach allows our model, i.e. Instruction to Facial Expression Transition (I2FET), to enhance the relationships between facial expression features and textual descriptions, thereby improving the latent representation for accurately generating target facial expressions and reducing the reliance solely on instructional text.
Then, facial expression trajectories are created using the target facial expressions, which are generated by the facial parameters of the source image using I2FET. Using this framework, one can generate a 3D facial expression avatar from a photograph of a face and create any facial transition based on a text prompt instruction that contains the initial and final facial expressions.
\\
\hspace*{0.3cm}The main contributions of our study are given as follows:
\vskip 0.5em
\begin{itemize} 
   \item We present an instruction-driven 3D facial expression generation and transition framework to generate transitions in facial expressions based on instructions, starting from a photograph of a face.
   \item We propose an Instruction-Driven Facial Expression Decomposer (IFED), designed to learn multimodal data and capture the correlation between facial expression features and textual descriptions.
   \item We propose an Instruction to Facial Expression Transition method (I2FET), which utilizes conditional variational autoencoders, integrated with IFED and a vertex reconstruction loss function, to enhance the semantic comprehension of the latent vectors. 
   \item Extensive evaluation of the proposed model on the CK+ and CelebV-HQ datasets suggests that it outperforms SOTA methods.
\end{itemize}

% \begin{figure}[ht!]
%     \centering
%     \includegraphics[scale=0.45]{figs/fet-problem.pdf}
%     \caption{Illustration of how to generate facial expression trajectories with the proposed FET, starting at an arbitrary expression of the input face and ending at the target expression instructed by text. DECA (Detailed Expression Capture and Animation) \cite{DECA:Siggraph2021} is used to encode shape $\phi_{s}$, expression $e_{s}$, and pose $\theta_{s}$ parameters of the input image. The instruction is given to the CLIP encoder \cite{radford2021learning} and then passed through a module, named T2FET decoder to generate corresponding expressions $e_{h}, e_{su}$, and poses $\theta_{h}, \theta_{su}$, which are given in the text. An additional face decoder is used to generate expressions $e_{n}$ and poses $\theta_{n}$ for the neutral expression. This is suitable for investigating facial behaviors, as depicted within the pleasure-arousal (P-A) space \cite{russel1997}, where the neutral expression serves as a transitional expression between two specific expressions. For smooth expressions, a linear interpolation function $\Psi$ is applied between anchor expressions to synthesize the transferring expressions, while the shape $\phi_{i}$ remains during this process since it allows preserving the identity of the subject in the input image and generating a long sequence with diverse expressions.} 
%     \label{fig:idea}
% \end{figure}

\section{Related Work}
\label{sec:relatedwork}
\subsection{Facial Expression Transition}
Generative Adversarial Networks (GANs) have been favored for generating facial expressions. For instance, Motion3DGAN \cite{Otberdout2022GeneratingC4}, an extension of MotionGAN \cite{otberdout2022sparse}, addresses the dynamics of 3D landmarks, whereas WGAN learns the distribution of 3D expression dynamics across the hypersphere space, sampled with a condition to generate landmark sequences. In this case, a neutral 3D face mesh, frame by frame, is deformed by a mesh decoder using the generated landmark sequences although the disadvantage of this work is that the start and end expressions have to be provided to generate a concatenated expression transition.     
On the other hand, a diffusion model is adopted for facial expression generation \cite{zou20234d} by modeling the input distribution. The deformation of the 3D mesh is guided by a specific landmark, and the original shape of the input facial mesh is considered to apply displacement to each vertex. 
In this work, MotionClip \cite{tevet2022motionclip}, which was originally designed for human motion synthesis, is compared with the task of generating the expression transition conditions given by the text.
EMOTE \cite{2023dan} utilizes emotion labels and various inputs to generate a talking-head avatar that maintains lip-sync from speech as well as facial expression. In particular, the facial motion prior, called FLINT (FLAME IN TIME), uses the expression and pose parameters of FLAME \cite{Li2017LearningAM} to represent facial motion sequences. 

Although several studies have demonstrated methods to produce transitions between facial expressions under diverse conditions, the generated faces are typically conditioned on the front face by neglecting the head pose variations. Inspired by previous works \cite{Otberdout2022GeneratingC4, zou20234d, 2023dan}, the present study proposes a model for facial expression transition that learns the expression and pose parameters. The additional merit of our model would be that it can easily be plugged into the models that generate the facial appearance.   
\subsection{Face Rendering}
% Recently, various studies have addressed the reconstruction of facial features, transitioning from 2D images to 3D.
%\subsubsection{Neural Textures} 
 In the face rendering area, 3D Morphable Model (3DMM) has been widely used for rendering realistic faces \cite{buehler2021varitex, leeexpgan2023, Khakhulin2022ROME, ma2023cvthead}.
Among these methods, VariTex \cite{buehler2021varitex} employs a neural texture obtained by a face texture decoder to sample with a UV map. Subsequently, this neural face feature image is merged with an additive feature image to serve as the input for a U-Net to generate the desired image. Exp-GAN \cite{leeexpgan2023} employs a neural volume generator and StyleGAN2 to generate images at various resolutions. This process combines different feature components, such as face features, depth images, and feature volumes. The facial feature is also obtained by sampling the neural texture using the UV map. However, this work is computationally intensive and time-consuming.
ROME \cite{Khakhulin2022ROME} can synthesize realistic face images by estimating a specific head mesh and neural texture from an input image, whereupon a neural rendering technique is adapted to render the rigged mesh. Next, the U-net-like neural rendering network, which accepts inputs such as neural texture and surface normals, is utilized to acquire both the target image and its corresponding mask.   
 Similarly, CVTHead \cite{ma2023cvthead} also estimates a neural texture from an input image, and yet, in this case, the neural texture is flattened to a feature vector form before being passed to transformers along with vertex tokens to output a vertex descriptor. Specifically, each of the vertices is considered a query token, and a transformer is employed to learn the canonical vertex feature from the source image. To generate the desired image and its mask, the feature descriptor of each vertex and depth are projected into image space before being processed by a U-Net.      
 
In the current study, we employ ROME and CVTHead to generate facial appearances by leveraging DECA for the extraction of facial expression coefficients. This streamlines the integration into a facial expression transition model and enhances the applicability of the proposed framework to practical scenarios.
\section{Method}
\label{sec:method}
\subsection{Problem Statement} 
We aim to develop a model to generate 3D facial expression with two inputs: a face image $I_s$ and a text instruction $t$. The facial expression transition model $\mathcal{T}_{w}(t, I_s)$, having weights $w$, is trained with a training set $\mathcal{X} = \{(t, (e_0, \theta_0), (e_1, \theta_1))^{(j)}\}_{j=1}^n$, where $e_0, e_1 \in R^{1 \times 50}$ denote expression vectors with 50 dimensions and $\theta_0, \theta_1 \in R^{1 \times 6}$ represent pose vectors with 6 dimensions corresponding to the first and next expressions in the facial expression sequence specified in instruction $t$, respectively. 
$n$ is the number of samples in the training dataset. Our network learns $\mathcal{T}_{w}(.)$ that approximates the data distribution $p(\mathcal{X})$ and then generates a facial expression coefficient sequence $\{(e_0, \theta_0), (e_1, \theta_1)\}$ with a given instruction $t$.
Next, a face rendering module $\mathcal{G}(I_s, \mathcal{S})$ is used to render a sequence of facial appearances $\mathcal{Y} = \{y_s^{(1)}, y^{(i)}, y_{e_0}^{(k)}, y^{(j)}, y_{e_1}^{(T)}\}$ , with inputs, consisting of the texture of source image $I_s$ and facial expression trajectories 
$S = \{s_s^{(1)}, s^{(i)}, s_{e_0}^{(k)}, s^{(j)}, s_{e_1}^{(T)}\}$, 
where $\{s_s^{(1)}, s_{e_0}^{(k)}, s_{e_1}^{(T)}\}$ are anchor facial expression sequences corresponding to the source image, and specific facial expressions, generated by $\mathcal{T}_{w}(.)$. 
Then, facial expression sequences $s^{(i)}$ and $s^{(j)}$ transitioning between these anchor facial expressions can be inferred by a linear interpolation function to ensure the smoothness of the sequence of facial expressions.  
\subsection{The Proposed Method}
\begin{figure}[ht!]
    \centering
    \includegraphics[scale=0.24]{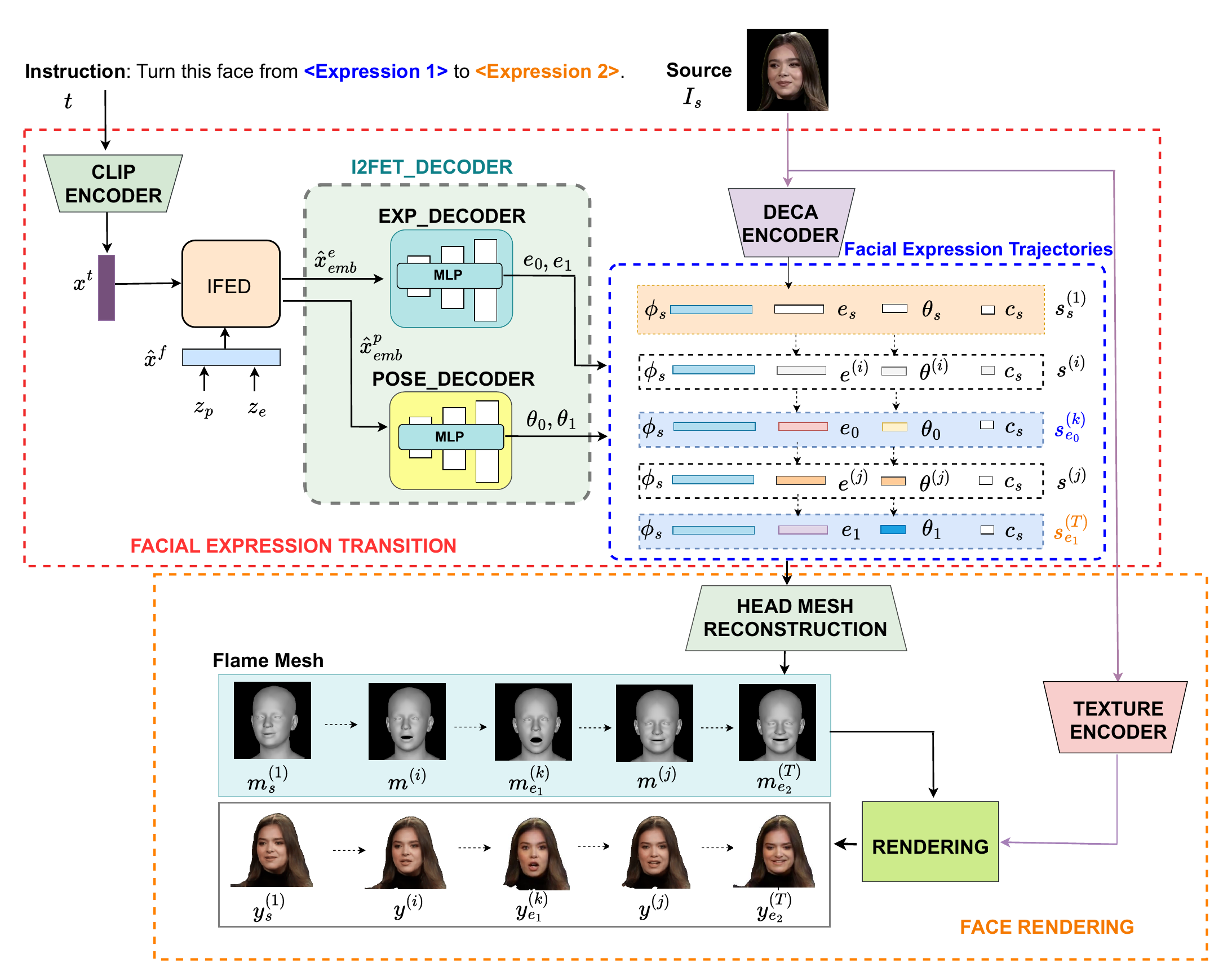}
    \caption{Overview of the proposed framework consisting of two major modules: Facial Expression Transition (FET) and Face Rendering (FR). First, textual vector $x_t$ is obtained from the pre-trained CLIP encoder for the FET module. Simultaneously, latent representations $z_{p}$ and $z_e$ are drawn from $\mathcal{N}(0, I)$. Then, the latent vectors are processed and concatenated to obtain $\hat{x}^f$. Afterward, the IFED module utilizes $x^t$ and $\hat{x}^f$ as inputs to create conditional feature vectors $\hat{x}_{emb}^e$, and $\hat{x}_{emb}^{p}$ for the pose and expression decoders. Subsequently, for smoothness of the facial expression sequence, the facial expression in the source image $(e_s, \theta_s)$ is subjected to linear interpolation and the specific facial expressions $(e_0, \theta_0), (e_1, \theta_1)$, generated by the I2FET decoders. 
    These sequences are then combined with the shape $\phi_s$ and camera $c_s$ parameters of the source image, obtained from DECA, to form facial expression trajectories $\{s_s^{(1)},s^{(i)},s^{(k)}_{e_0},s^{(j)}, s_{e_1}^{(T)}\}$. For FR, head mesh reconstruction produces a flame mesh sequence $\{m_s^{(1)},m^{(i)},m^{(k)}_{e_0},m^{(j)}, m_{e_1}^{(T)}\}$ aligned with the facial expression trajectories and the texture encoder extracts the facial appearance from the source image. Finally, the rendering module generates the facial appearances $\{y_s^{(1)},y^{(i)},y^{(k)}_{e_0},y^{(j)}, y_{e_1}^{(T)}\}$  within the facial expression trajectories using the flame mesh sequence and the extracted facial appearance.
    }
    \label{fig:proposedmethod}
\end{figure}

Figure \ref{fig:proposedmethod} provides an overview of the proposed framework whereby one can control 3D facial expressions of the monocular image $I_s$ following an input instruction $t$. 
The framework consists of two major modules: (1) Facial Expression Transition (FET), $\mathcal{T}_{w}(t, I_s)$, for generating diverse expression trajectories; (2) Face Rendering (FR), $\mathcal{G}(I_s, S)$, for rendering facial appearances based on the texture of the source image $I_s$ and the expression trajectories $S$ generated by the FET module.  
\subsubsection{Facial Expression Transition}
This module is designed to generate diverse expression trajectories that correspond to the facial parameters of the source face image $I_s$ and the parameters of specific facial expressions, described in an instruction $t$. The module has two primary branches: the first receives an instruction to generate the corresponding sequence of facial expression based on instruction $t$, whereas the second takes a 2D face image to encode its facial coefficients and create the facial expression trajectories. 
%%%%%%%%%% FIGURES %%%%%%%%%%
\begin{figure}[ht!]
    \centering
    \includegraphics[scale=0.25]{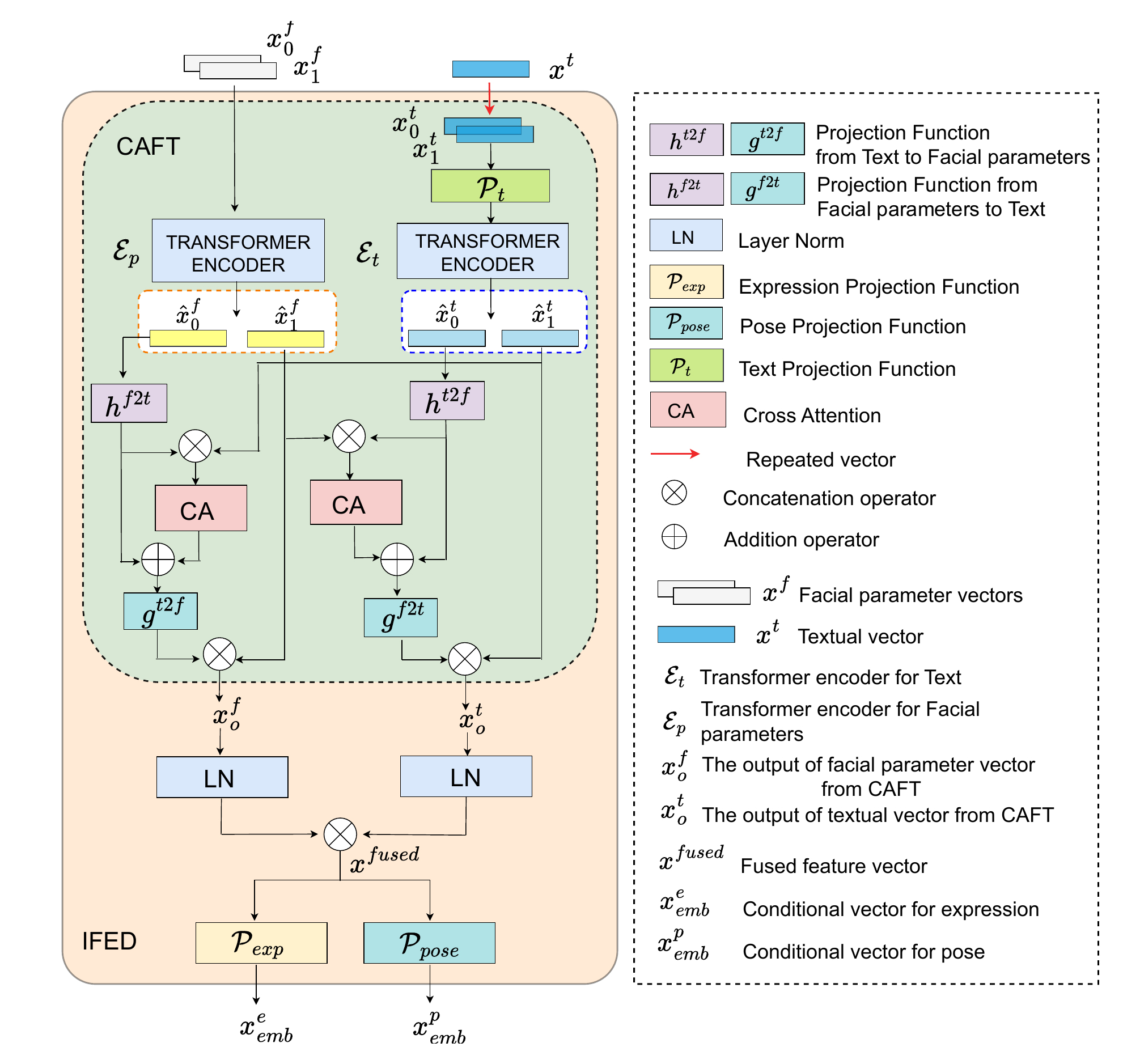}
    \caption{Flow diagram of IFED (Instruction-Driven Facial Expression Decomposer). It consists of CAFT and the decomposition of parameters. The major function of CAFT is to introduce cross-attention between the facial parameters and text instruction. The two outputs from CAFT pass through layer normalization and are combined to create the fused feature vector $x^{fused}$. Then, pose projection function $\mathcal{P}_{p}$ and expression projection function $\mathcal{P}_{e}$ are used to decompose $x^{fused}$ into two conditional vectors for facial expression ($x_{emb}^e$) and pose ($x_{emb}^{p}$).}
    \label{fig:ftd}
\end{figure}
\begin{figure*}[ht!]
    \centering
    \includegraphics[scale=0.31]{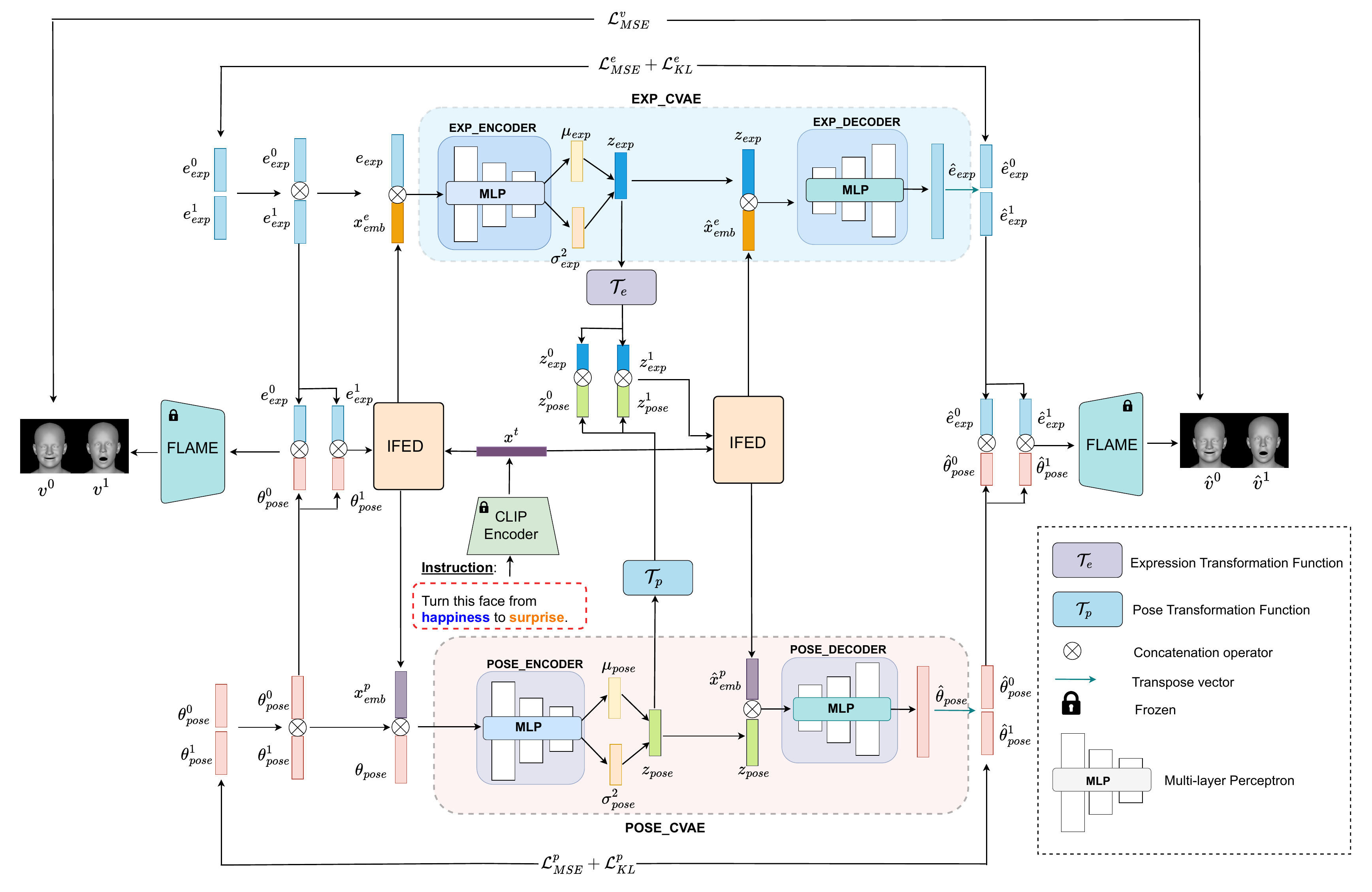}
    \caption{Detailed architecture of the I2FET (Instruction to Facial Expression Transition) method. Given an instruction and a sequence of facial parameters, including pose and expression, the IFED module utilizes the facial parameters and textual description extracted by CLIP to produce the corresponding conditional feature vectors for pose $x_{emb}^{p}$ and expression $x_{emb}^e$, respectively. The pose and expression encoders then map this sequence of facial parameters and conditional feature vectors to latent representations corresponding to pose and expression. These latent vectors, along with the textual description, are passed through the IFED module to generate the conditional feature vectors $\hat{x}_{emb}^e$ and $\hat{x}_{emb}^{p}$. Following this, the expression and pose decoders reconstruct expression $\hat{e}$ and pose $\hat{\theta}$ parameters using these latent representations and the conditional feature vectors $\hat{x}_{emb}^e$ and $\hat{x}_{emb}^{p}$, respectively. Finally, the pre-trained FLAME decoder reconstructs the vertex coordinates $\hat{v}$ based on $\hat{e}$ and $\hat{\theta}$.}
    \label{fig:t2fet}
\end{figure*}
%%%%%%%%%%

In the former, the pre-trained CLIP encoder \cite{radford2021learning} extracts text description $x^t \in R^{77 \times 768}$ according to the given instruction $t$. Then, the I2FET model generates a sequence of expression and pose coefficients for the facial expression trajectories. 
%%%%% T2FET 
\\
\hspace*{10pt}\textit{Instruction-driven Facial Expression Decomposer}:
A dual-branch vision transformer \cite{Chen2021CrossViTCM} is adopted for extracting multi-scale features and fusing cross-attention information at different scales because it is efficient for the image classification task. Here, the Cross-Attention for Facial Expression Parameter and Text (CAFT) is used to learn the relationship between instruction and facial expression parameters by enhancing the understanding of multimodal data and refining the latent representation for the Instruction to Facial Expression Transition model. The proposed IFED module is illustrated in Fig. \ref{fig:ftd}. In particular, we directly incorporate facial expression parameters and textual descriptions into CAFT instead of adding class tokens as in \cite{Chen2021CrossViTCM}. With respect to the transformer encoders, the first branch learns the correlation between the facial expression parameters, while the second branch focuses on learning the textual descriptions. To begin with, each vector representing facial expression parameters comprises two elements: expression and pose parameters, denoted as $x^f=\{e,\theta\} \in R^{m \times 56}$, where $m$ denotes the number of expressions in a given instruction. As \cite{LinCGF2023, 2023dan}, we only utilize jaw pose $\theta^{jaw}$ and expression $e$ to convey facial expression information. Therefore, we rewrite $x^f = \{e, \theta^{jaw} \} \in R^{m \times 53}$ as the input of the first branch. Then, the transformer encoder is utilized as
\begin{equation}
\begin{aligned}
y^f = x^f + MSA(LN(x^f)) \\
\hat{x}^f = y^f + FFN(LN(y^f))
\end{aligned}
\end{equation}
where $\hat{x}^f$ denotes the output of the transformer encoder in the facial expression parameters branch $\mathcal{E}_P$. $LN(.)$, $MSA(.)$, and $FFN(.)$ are the layer norm, multi-head self-attention, and feed-forward network, respectively. Likewise, in the text branch, we replicate $x^t$ to create a vector $\{x\}^t_{i=\{0,1\}}$ for computational purposes. Subsequently, the input for the transformer encoder is generated using a text projection function $P_t(.)$ on $\{x\}^t_{i=\{0,1\}}$. As a result, the textual description $\hat{x}^t$ is obtained through the transformer encoder $\mathcal{E}_t$.
Both branches are then fused by cross-attention to capture the relationship between the facial expression parameters and textual descriptions. The fusion process can be expressed as
\begin{equation}
    \begin{aligned}
        y_c^f = h^{f2t}(\hat{x}_0^f), \quad y_c^t = h^{t2f}(\hat{x}_0^t), \\
        x_o^f = g^{t2f}([CA(y^f_c \otimes x_1^t) + y_c^f]) \otimes \hat{x}_1^f \\
        x_o^t = g^{f2t}([CA(y_c^t \otimes x_1^f) + y_c^t]) \otimes \hat{x}_1^t
    \end{aligned}
\end{equation}
where $h^{t2f}, h^{f2t}, g^{t2f}$, and $g^{f2t}$ are the projection function and back-projection function between the vectors that represent the text and facial expression parameters, $x_o^f \in R^{m \times 53}$ and $x_o^t \in R^{m \times 768}$ are the outputs of CAFT, $CA$ denotes cross-attention and $\otimes$ denotes the concatenation operator.  
Next, the fused feature vector is obtained by concatenating $x_o^f$ and $x_o^t$, after passing through $LN(.)$, as follows: 
\begin{equation}
    x^{fused} = LN(x_o^t) \otimes LN(x_o^f)
\end{equation}
Finally, the projection functions for pose $\mathcal{P}_{p}(.)$ and expression $\mathcal{P}_{e}(.)$ are applied to the fused feature vector as follows:
\begin{equation}
  \begin{aligned}
    x^e_{emb} = \mathcal{P}_{e}(x^{fused}) \\
    x^{p}_{emb} = \mathcal{P}_{p}(x^{fused}) 
  \end{aligned}
\end{equation}
where $x^e_{emb} \in R^{m \times 50}$ and $x^{p}_{emb} \in R^{m \times 6}$ are the conditional feature vectors for pose and expression, respectively. In this work, $\mathcal{P}_{p}(.)$ and $\mathcal{P}_{e}(.)$ are designed as linear transformation functions to map the feature $x^{fused}$ to the corresponding pose and expression feature vectors.
\\
\hspace*{10pt}\textit{Instruction to Facial Expression Transition}: 
The proposed I2FET method is based on a conditional variational autoencoder architecture. The method includes encoders $\mathcal{E}_e^t (.)$ and $\mathcal{E}_p^t(.)$, as well as decoders $\mathcal{D}^t_e(.)$ and $\mathcal{D}^t_p(.)$, which are defined using three Multi-Layer Perceptrons (MLPs). These components are conditioned on feature vectors that represent expression and pose information, respectively. An overview of the proposed I2FET architecture is presented in Fig.\ref{fig:t2fet}.

Here, the encoders are defined as follows: 

\begin{equation}
\mathcal{E}_e^t (e \otimes x^{e}_{emb}) \longrightarrow (\mu_e, \sigma_e) 
\end{equation}
 and 
 \begin{equation}
 \mathcal{E}_p^t (\theta \otimes x^{p}_{emb}) \longrightarrow (\mu_p, \sigma_p)
 \end{equation}
 where $\otimes$ denotes the concatenation operator and $x^e_{emb}$ and $x^{p}_{emb}$ are conditional feature vectors for expression and pose, respectively. For the encoder stage, $x^e_{emb}$ and $x^{p}_{emb}$ are created by textual description $x_t$ and the facial expression parameters $(e, \theta)$. $\mu_e$ and $\mu_p$ denote the mean and standard deviation of the expression distribution, whereas $\sigma_e$, and $\sigma_p$ are the mean and standard deviation of the pose distribution. In this study,   
 the conditional feature vectors $\hat{x}^e_{emb}$ and $\hat{x}^{p}_{emb}$ in the decoder are computed by textual description $x_t$ and the latent vectors $\hat{z}_p$ and $\hat{z}_{e}$ representing pose and expression, respectively. These latent vectors are calculated as follows:  
 \begin{equation}
  \begin{aligned}
  \tilde{z}_p = \sigma_p*z_p + \mu_p \\
  \hat{z}_p = \mathcal{T}_{p}(\tilde{z}_p) \\
  \tilde{z}_e = \sigma_e*z_e + \mu_e \\
  \hat{z}_e = \mathcal{T}_{e}(\tilde{z}_e) \\
   \end{aligned}
 \end{equation}
   where $z_p$ and $z_e$ are the latent vectors sampled from $\mathcal{N}(0, I)$.
   In this case, we utilize linear transformation functions $\mathcal{T}_{p}(.)$ and $\mathcal{T}_e(.)$ to transform $\tilde{z}_p$ and $\tilde{z}_e \in R^{m \times 16}$ into $\hat{z}_p \in R^{m \times 6}$ and $\hat{z}_e \in R^{m \times 50}$, respectively, similar to the expression and pose dimensions. 
   Thus, the latent vectors $\tilde{z}_e$ and $\tilde{z}_p$ can be used as input for the IFED module. Then, the latent vectors $\hat{z}_e$ and $\hat{z}_p$ are concatenated together, and along with the text vector $x^t$, they are passed through the IFED module to obtain the conditional feature vectors $\hat{x}_{emb}^e$ and $\hat{x}_{emb}^p$ as     
   \begin{equation}
   \begin{aligned}
   \hat{x}_{emb}^{e}, \hat{x}_{emb}^{p} = IFED(\hat{z}_p \otimes \hat{z}_e, x^t)  
   \end{aligned}
   \end{equation}

 This approach enhances the semantic representation of latent vectors by refining the conditional feature vectors for decoders based on the learned latent vectors throughout the training process.
 The decoders are defined as follows
  \begin{equation}
  \mathcal{D}^t_p( \tilde{z}_p \otimes \hat{x}^{p}_{emb}) \longrightarrow (\hat{\theta}_0, \hat{\theta}_1)
  \end{equation}
  and
  \begin{equation}
  \mathcal{D}^t_e(\tilde{z}_e \otimes \hat{x}^{e}_{emb}) \longrightarrow (\hat{e}_0, \hat{e}_1)
  \end{equation}

The parameters of the I2FET model are optimized by minimizing the reconstruction loss function $\mathcal{L}_{MSE}(.,.)$ between the ground truth coefficients $e$, $\theta$, and the predicted coefficients $\hat{e}, \hat{\theta}$. In addition, the Kullback-Leibler divergence loss function is utilized to minimize the distance between the probability distribution of the ground truth and that of the predicted values.
\begin{equation}
  \begin{aligned}
    \mathcal{L}_{e} = \mathcal{L}_{MSE}(e,\hat{e}) \\
    + 0.5 \left[ -\sum_i(log \: \sigma_i^2 + 1)  + \sum_i\sigma^2_i  
    + \sum_i \mu_i^2  \right] 
  \end{aligned}
\end{equation}
\begin{equation}
  \begin{aligned}
    \mathcal{L}_{p} = \mathcal{L}_{MSE}(\theta,\hat{\theta})  \\ +0.5 \left[ -\sum_i(log \: \sigma_i^2 + 1) + \sum_i\sigma^2_i + \sum_i \mu_i^2 \right] 
  \end{aligned}
\end{equation}
where $\mathcal{L}_{e}$ and $\mathcal{L}_{p}$ denote the expression and pose loss function, respectively.

%%%%%% FLAME %%%%
Furthermore, we utilize the pretrained FLAME \cite{Li2017LearningAM}, a parametric model of a 3D head that is integrated into DECA \cite{DECA:Siggraph2021}, which has $N=5023$ base vertices $V_{b} \in \mathcal{R}^{N\times 3}$ and two sets of $M$ and $K$ basis vectors that encode shape blendshapes $\mathcal{S} \in R^{N \times 3 \times M}$, and expression blendshapes $B \in R^{N \times 3 \times K}$. The basic vectors are first blended and then Linear Blend Skinning (LBS) $\mathcal{W}(.)$ is applied to rotate the vertices following the pose $\theta$.  
The final reconstruction in world coordinates can be computed as follows
\begin{equation}
    \mathcal{V}(\phi, e, \theta) = \mathcal{W}(V_b + \mathcal{S}\phi + \mathcal{B}e, \theta) 
    \label{eq:flame_vetex}
\end{equation}
where $\phi$, $e$, and $\theta$ denote the identity shape, facial expression, and head pose parameters, respectively.  
%{\color{red}{
In this case, FLAME is used to reconstruct vertices to improve the performance of I2FET. 
\begin{equation}
    \mathcal{L}_{v} = \mathcal{L}_{MSE}(v, \hat{v})
\end{equation}
where the vertex coordinates $v$ and $\hat{v}$ are computed by feeding the ground truth $\{\theta, e, \phi\}$ and the reconstructed $\{\hat{\theta}, \hat{e}, \phi\}$ through FLAME.

Finally, the total loss function is given as 
\begin{equation}
\mathcal{L}_{total} = \mathcal{L}_{e} + \mathcal{L}_{p} + \mathcal{L}_{v} 
\end{equation}
%}}
%%%%%%%%%%%%%%%%%%%%%%%%%%%%%%%%%%%%
For the second branch, the pre-trained DECA \cite{DECA:Siggraph2021} is used to encode the facial coefficients of the source face $I_s$ including shape $\phi_s \in R^{1 \times 100}$, expression $e_s \in R^{1 \times 50}$, pose $\theta_s \in R^{1 \times 6}$ and camera $c_s \in R^{1 \times 3}$. 
%{\color{red}{
Following this step, $\{ \phi_s, e_s, \theta_s, c_s\}$ is integrated with the sequence of facial expression coefficients $\{(e_0, \theta_0), (e_1, \theta_1)\}$, generated by the first branch to generate the anchor facial expressions $\{s^{(1)}_s, s^{(k)}_{e_0}, s^{(T)}_{e_1}\}$.
%}
Then, a linear interpolation function $\Psi(.)$ is applied to smoothen the transition between the two anchored facial expressions. The $\Psi(s^{(l)}, s^{(n)})$ function for interpolation between facial expression sequences $s^{(l)}$ and $s^{(n)}$ can be defined as 
\begin{equation}
\begin{aligned}
e^{(k)} = \delta*e^{(l)} + (1 - \delta)*e^{(n)}    \\
\theta^{(k)} = \delta*\theta^{(l)} + (1 - \delta)*\theta^{(n)} \\
\end{aligned}
\end{equation}
where $0 \leq \delta \leq 1$ is a linear coefficient and $\{e^{(l)}, \theta^{(l)} \} \in s^{(l)}$. This allows our framework to ensure temporal consistency when the motion changes are sampled uniformly.
 
\subsubsection{Face Rendering}
This module renders the facial appearance for the facial expression trajectories, generated by FET. In this work, the face rendering approach is based on the facial parameters of FLAME for integration with FET.
\\
\hspace*{10pt}\textit{Head Mesh Reconstruction:} 

It generates the vertex coordinates corresponding to the facial expressions using Eq.\ref{eq:flame_vetex} by using the sequence of facial expression transitions as the input. Then, the target vertex coordinates are obtained by combining these expressions with the deformation of hair and shoulder regions $f_H(.)$ from $I_s$ as follows 
\begin{equation}
\mathcal{M}^t = \mathcal{V}(\phi, e, \theta) + f_H (I_s)
\end{equation}
where $f_H(.)$ is the pre-trained linear deformation model \cite{Khakhulin2022ROME} that is used to refine the vertices locations. 
\\
\hspace*{10pt}\textit{Rendering}:

Following the head mesh reconstruction step, facial appearances can be rendered from the reconstructed head vertices $\mathcal{M} = \{m_s^{(1)},m^{(i)}, m_{e_0}^{(k)}, m^{(j)}, m_{e_1}^{(T)}\}$ and the extracted neutral texture of the source image $I_s$ by the texture decoder $\mathcal{E}_{tex}(I_s)$.
The present work explores a face-rendering model suitable for facial expression transition. Therefore, we utilize pre-trained facial rendering models, known as state-of-the-art models \cite{Khakhulin2022ROME, ma2023cvthead}, capable of rendering facial appearances based on the facial parameters calculated with FLAME. 
Subsequently, the sequence of facial appearances $\mathcal{Y} = \{y_s^{(1)}, y^{(i)}, y_{e_0}^{(k)}, y^{(j)}, y_{e_1}^{(T)}\}$ is generated as the output of the proposed framework.      

\begin{figure}[ht!]
%\belowcaptionskip = -15pt
\centering
%%%%%%%%%% CK+ %%%%%%%%
\begin{subfigure}{0.47\linewidth}
    \includegraphics[scale=0.45]{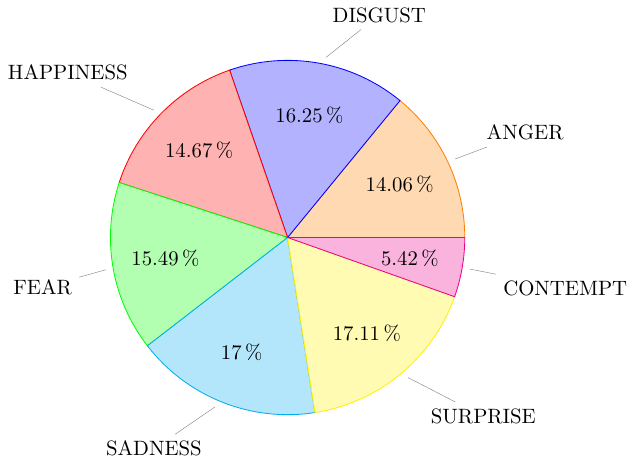}
    \caption{ }
    \label{fig:tfeg-ck}
\end{subfigure}
%%%%%%%%%% CelebV-HQ %%%%%%%%
\begin{subfigure}{0.41\linewidth}
    \includegraphics[scale=0.45]{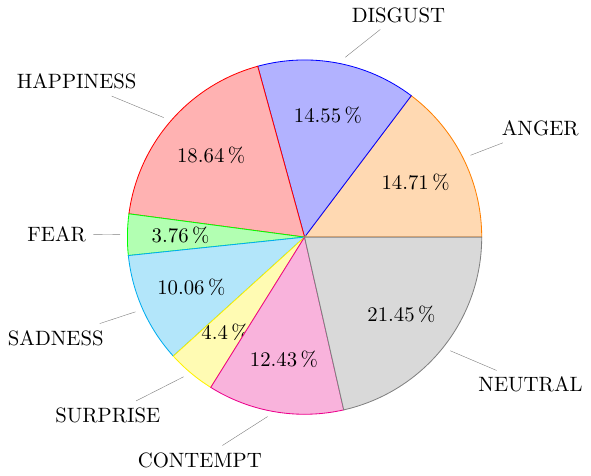}
    \caption{ }
    \label{fig:tfeg-celebv-hq}
\end{subfigure}
% \hfill
\begin{subfigure}[b]{0.4\linewidth}
    \includegraphics[scale=0.35]{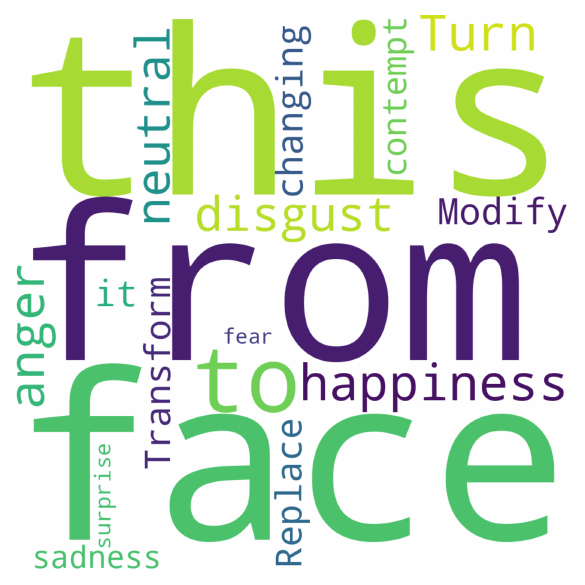}
    \caption{ }
    \label{fig:tfeg-celebv-hq_word}
\end{subfigure}
 \caption{Statistics of the facial expressions utilized in this study for the CK+ (a) and CelebV-HQ (b) datasets, respectively, and a collection of sampled words extracted from the text instructions (c).}
  \label{fig:short}
\end{figure}
%%%%%%%%%%%%%%%%%%%%%%%%%%%%%%%%%%%%%%%%%%%%%%%%%%%%%%%%%%%%%%%%%%%
\section{Experiments}
\label{sec:experiment}
\subsection{Dataset}
\begin{table}[ht]
    \centering
    % \caption{
    % $N_1,N_2  \in \{$ happiness, disgust, anger, fear, surprise, contempt, sadness$\}$ in CK+ dataset and $N_1,N_2 \in \{$ happiness, disgust, anger, fear, surprise, contempt, sadness, neutral$\}$ in CelebV-HQ dataset.
    % }
     \caption{
    $N_1,N_2  \in \{$ HAPPINESS, DISGUST, ANGER, FEAR, SURPRISE, CONTEMPT, SADNESS$\}$ IN THE CK+ DATASET AND $N_1,N_2 \in \{$ HAPPINESS, DISGUST, ANGER, FEAR, SURPRISE, CONTEMPT, SADNESS, NEUTRAL$\}$ IN THE CELEBV-HQ DATASET.
    }
    \label{tab:sen_template}
    \begin{tabular}{l|l}
     \hline
         No & Instruction/ Prompts  \\
         \hline
         1 & Turn this face from [$N_1$] to [$N_2$].  \\
         2 & Change this face from [$N_1$] to [$N_2$]. \\
         3 & Transform this face from [$N_1$] to [$N_2$]. \\
         4 & Modify this face, changing it from [$N_1$] \\
          & to [$N_2$]. \\
         5 & Replace this face from [$N_1$ to [$N_2$]. \\
         \hline
    \end{tabular}
\end{table}

\subsubsection{CK+ dataset} 
 % TFEG-CK+ dataset is developed based on the CK+ dataset by adding text instructions for facial expression generation and transition. 
 The CK+ dataset was supplemented with text instructions for facial expression generation and transition. The original CK+ dataset \cite{ckdb} includes 117 subjects and 7 emotions, such as \textit{happiness, disgust, anger, fear, surprise, contempt, and sadness}. The instructions are created to describe the transitions between these facial expressions. 
 To extract the facial parameters and the shape image corresponding to the facial expression in the image, we utilize the pre-trained DECA model. For each sample, a tuple is created with an instruction, the target shape image, and facial parameters that correspond to the same subject. The CK+ dataset, in this study, contains a total of 26,352 samples, and the composition of this dataset in terms of the seven facial expressions is shown in Fig. \ref{fig:tfeg-ck}. 
\subsubsection{CelebV-HQ dataset} 
Similar to the CK+ case, text instructions were added to the CelebV-HQ dataset for our experiments\cite{zhu2022celebvhq}.
The original CelebV-HQ dataset included 15,653 subjects and 8 expressions of \textit{neutral, happiness, disgust, anger, fear, surprise, contempt, and sadness}. The images in the CelebV-HQ dataset were spontaneously collected and presented many challenges such as highly non-frontal head poses, blur, and brightness. This problem was overcome by using a head pose algorithm \cite{Li9693249} to estimate parameters such as the \textit{yaw, roll, and pitch}, of the head pose ranging from -15 to 15 degrees. Given that the subjects were chosen indiscriminately, we select those with more than two facial expressions. Then, the pre-trained DECA was applied to extract the corresponding shape images and facial parameters. The CelebV-HQ dataset adopted in this study includes 614 subjects and 28,335 samples, and Fig. \ref{fig:tfeg-celebv-hq} shows a statistical breakdown of the eight facial expressions. \\ 
%%%%%%%%
\hspace*{10pt} Table \ref{tab:sen_template} lists the template sentences that are used to generate guided instructions for facial expression transitions. We generated five instructions for each sample and depicted the transition from expression $N_1$ to expression $N_2$. Subsequently, one presentation sentence is randomly chosen from these instructions. Fig. \ref{fig:tfeg-celebv-hq_word} illustrates some sampled words used in the text instructions.

\begin{figure}[ht!]
\belowcaptionskip = -15pt
\centering
\includegraphics[scale=0.15]{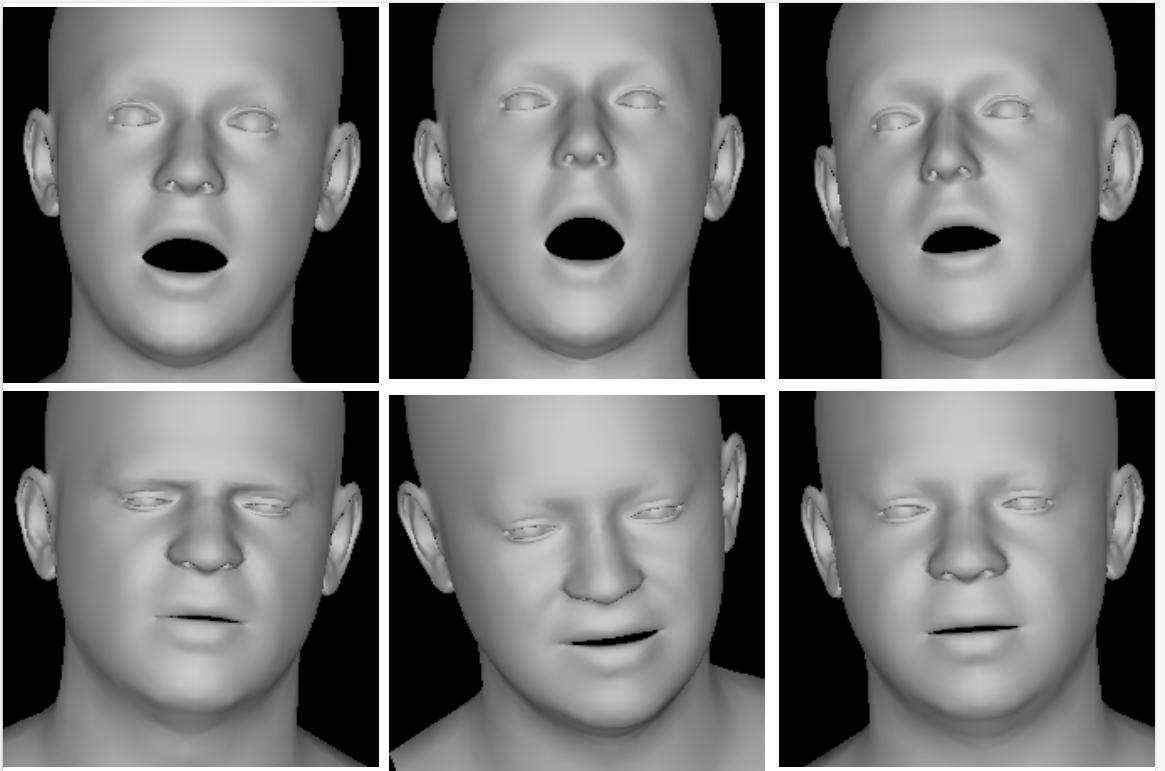}
    \caption{The shapes of facial expression are generated using images from the CK+ dataset. The top and bottom rows illustrate surprised and angry expressions with various head poses.}
    \label{fig:pose_tfeg-ck}
\end{figure}

%%%%%%%%%%%%%%%%%%%%%%%%%%%%%%%%%%%%%%%%%
\subsection{Implementation}
The I2FET model was trained for 200 epochs with a batch size of 128 and a learning rate of $8e^{-4}$. The Adam optimizer \cite{Kingma2014AdamAM} was adopted to optimize the model. For the experiments, 10\% of the dataset was allocated to the testing set, whereas the remaining portion was used for training. Then, a validation set was drawn from 10\% of the training set. In this work, we repeated each experiment ten times and used the average results to evaluate the overall performance of the proposed I2FET.
Because our task is new and a specific method that would allow a fair comparison is not available, we chose MotionClip \cite{tevet2022motionclip} as a baseline. For the face rendering experiment, the configuration by \cite{Khakhulin2022ROME,ma2023cvthead} was used.
%%%%%%%%%%%%%%%%%%%%%%%%%%%%%%%%%%%%%%%%%
% %%%%%%%%%%%%%%%%%%%% FACIAL EXPRESSION CLASSIFICATION %%%%%%%%%%%%%%%%%%%%
\subsection{Classification of Facial Expressions}
\label{sec:FEC}
\hspace{10pt} In this study, it was important to employ an appropriate classifier to evaluate the method that was used to generate facial expressions as prior studies \cite{Otberdout2022GeneratingC4, zou20234d}. Given that the CK+ and CelebV-HQ datasets had skewed distributions, the performance for minor classes could be affected by the facial expression classifiers trained on these imbalanced datasets, in which the results would be biased towards the major class. Moreover, the effectiveness of the classification model could also be affected by head movements, which were commonly encountered as shown in Fig. \ref{fig:pose_tfeg-ck}.

These problems were addressed by conducting an experiment to identify an appropriate classifier that can handle not only imbalanced datasets but also invariant rotation transformations. ResNet-101 \cite{he2016residual}, MEK \cite{zhang2023leave}, and ResNet-101 integrated with Reweighted Focal Loss (RFL) \cite{Cui2019ClassBalancedLB} were utilized to evaluate the performance of facial expression classification. 
For this purpose, augmented datasets were created by rotating the original CK+ and CelebV-HQ datasets at angles of $\pm 10^o, \pm 15^o$, and $\pm 30^o$. Then, the three networks are trained using the original datasets and evaluated on all datasets. The Accuracy was employed to assess whether samples had been correctly classified across all classes, while G-mean was used to evaluate the ability of the model to effectively recognize minor classes that are predicted by the facial expression classification model.  
The supplementary data in the ablation study in Section \ref{sec:ablstu} suggests that MEK can address imbalanced datasets but is sensitive to rotational transformations. ResNet101 demonstrates positive results with rotation transformations but performs worse on datasets that are highly imbalanced. ResNet101 integrated with RFL outperforms ResNet101 and MEK in both scenarios. As a result, ResNet101 with RFL was selected as our classifier for evaluating the ability of our models to generate facial expressions.
%%%%%%%%%%%%%%%%%%%% RESULT %%%%%%%%%%%%%%%%%%%%%%%
\subsection{Results}
\label{sec:result}
%\vskip 1em
%%%%%%%%%%%%%%%%%%%%%%%%%%%%%%%%%%%%%%%%%%%%%%%%%%%%%%%%%%%%%%%
%%%%%%%%%%%%%%%% t-SNE %%%%%%
\begin{figure}[ht!]
%\centering
% %%%%%%%%%% CK+ %%%%%%%%
\begin{subfigure}{0.1\linewidth}
    \includegraphics[scale=0.16]{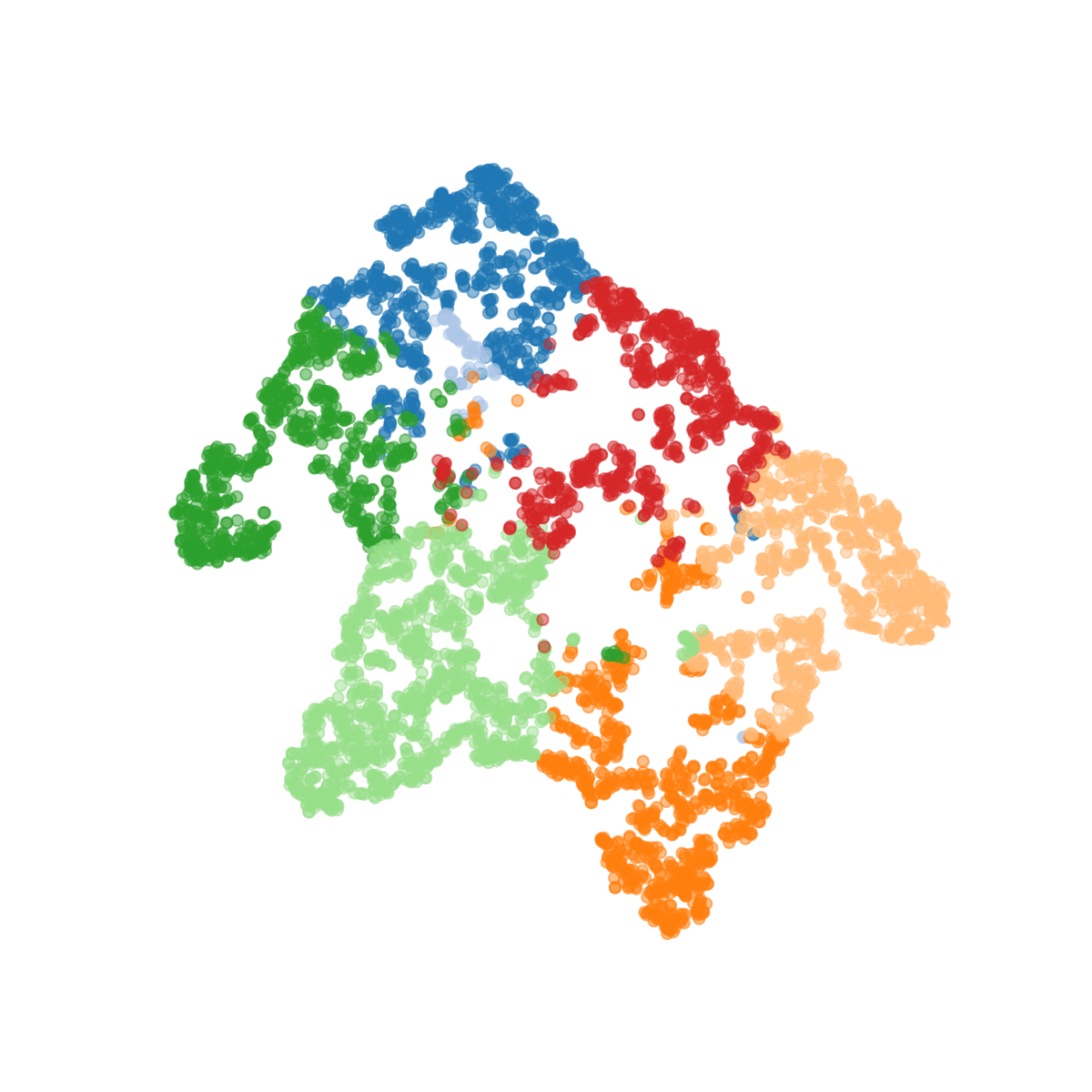}
    \caption{}
    \label{fig:tSNE-motionclip-ck}
\end{subfigure} 
 \hspace{4\baselineskip}
 \begin{subfigure}{0.2\linewidth}
     \includegraphics[scale=0.16]{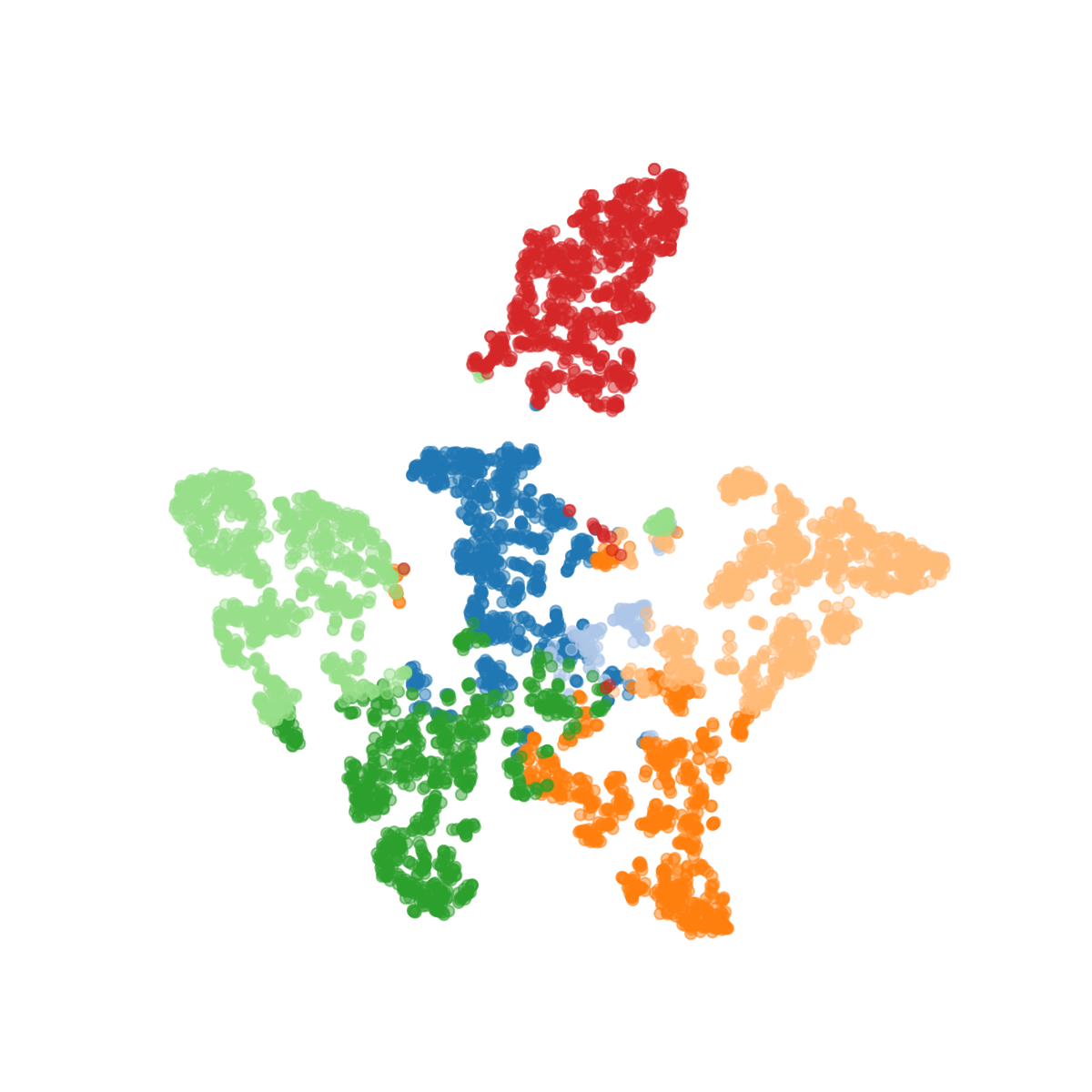}
     \caption{}
     \label{fig:tSNE-t2ft-ck}
\end{subfigure}
 \hspace{2.4\baselineskip}
\begin{subfigure}{0.2\linewidth}
    \includegraphics[scale=0.1]{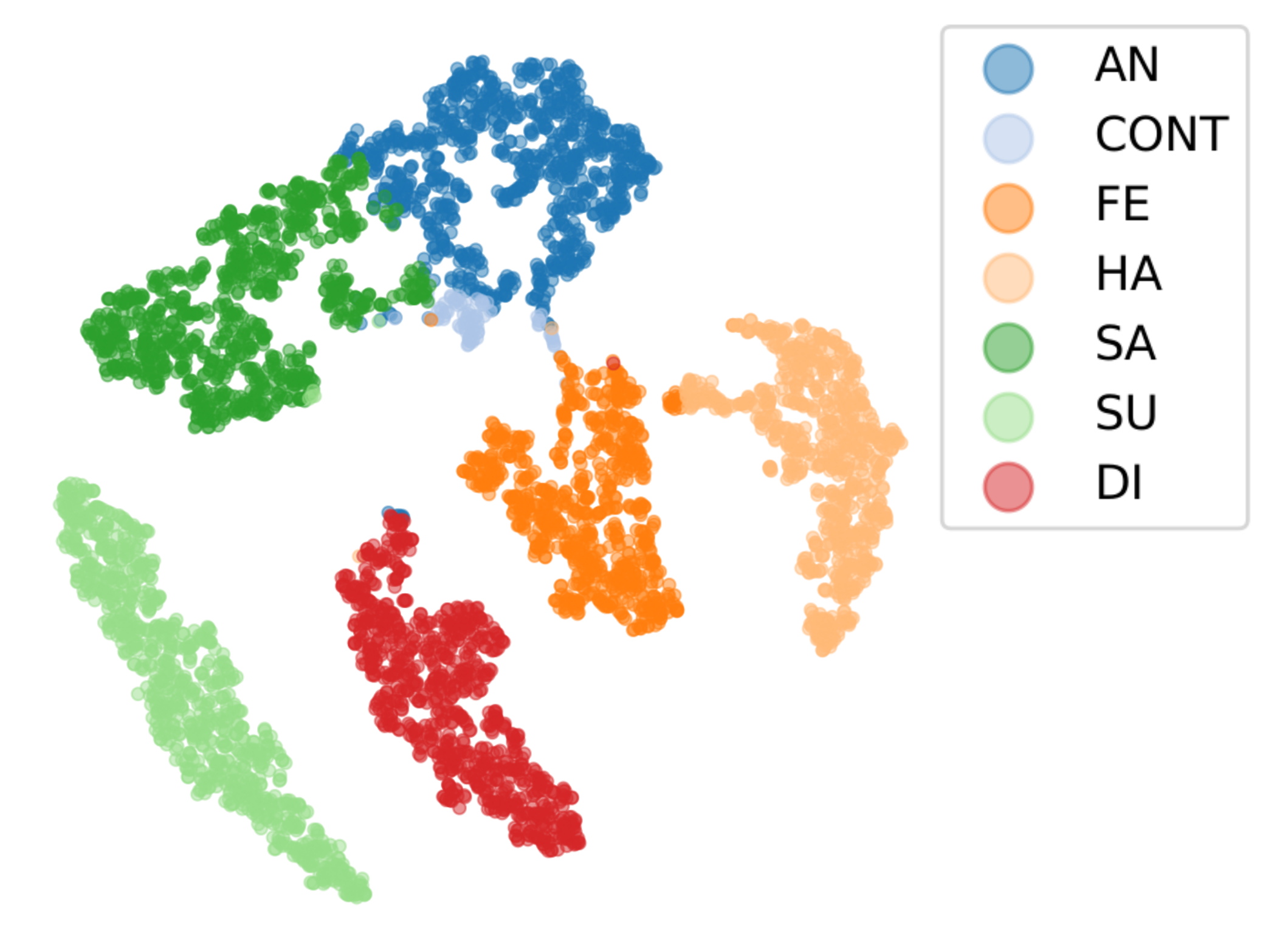}
    \caption{}
    \label{fig:tSNE-ck}
\end{subfigure}
 \caption{Visualization of the learned latent space for seven facial expressions by (a) MotionClip, (b) I2FET without IFED, and (c) I2FET with IFED, respectively, using t-SNE.}
 \label{fig:tSNE}
\end{figure}

%%%%%%%%%%%%%%%%%%%%%%%%%%%%%%%%%%%%%%%%%%%%%%
\begin{figure}[ht!]
%\centering
%%%%%%%%%% Motion clip %%%%%%%%
\begin{subfigure}{0.48\linewidth}
    \includegraphics[scale=0.251]{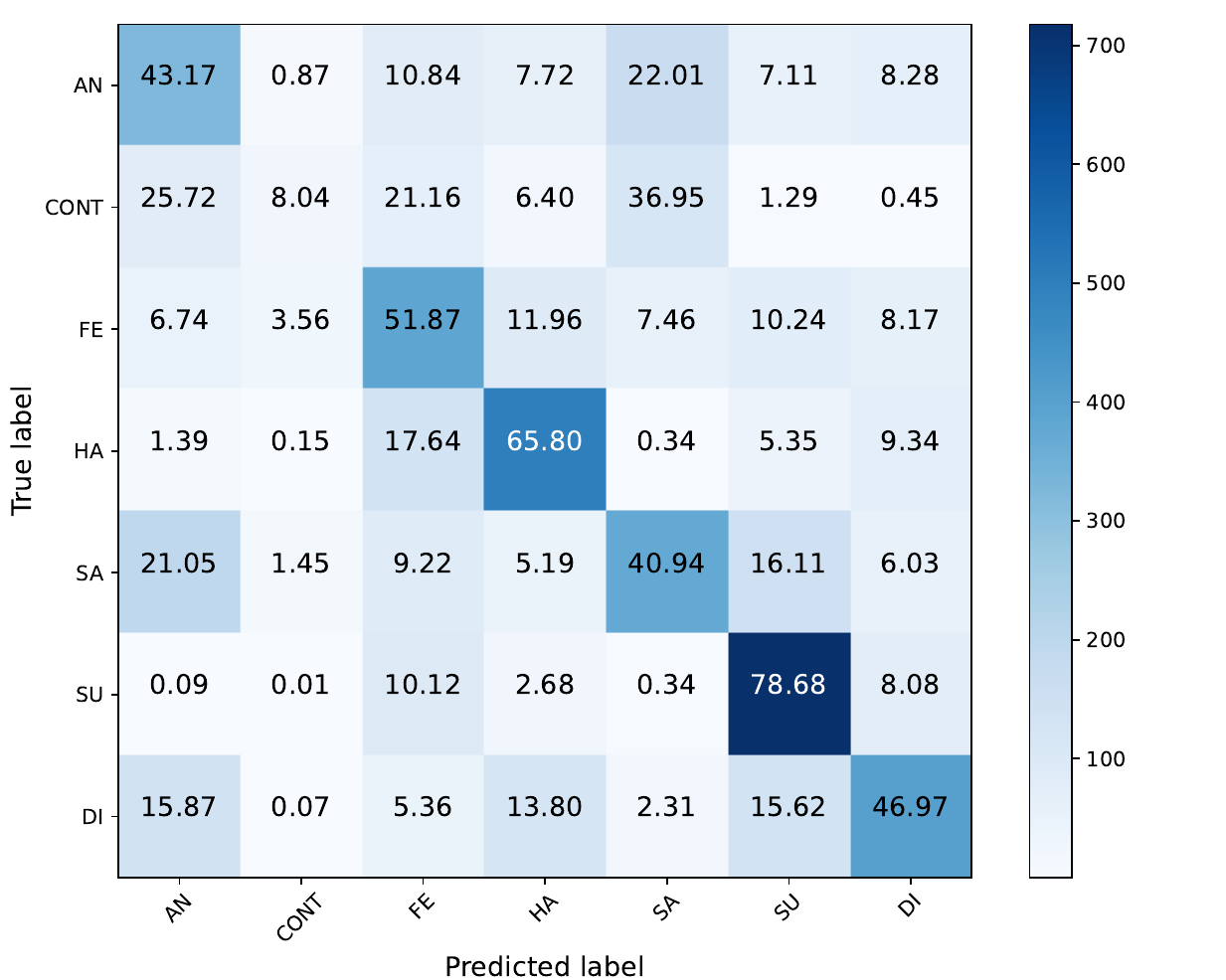}
    \caption{ }
    \label{fig:cfm-ck-motionclip}
\end{subfigure}
%%%%%%%%%% T2FT + TFED %%%%%%%%
\begin{subfigure}{0.45\linewidth}
    \includegraphics[scale=0.254]{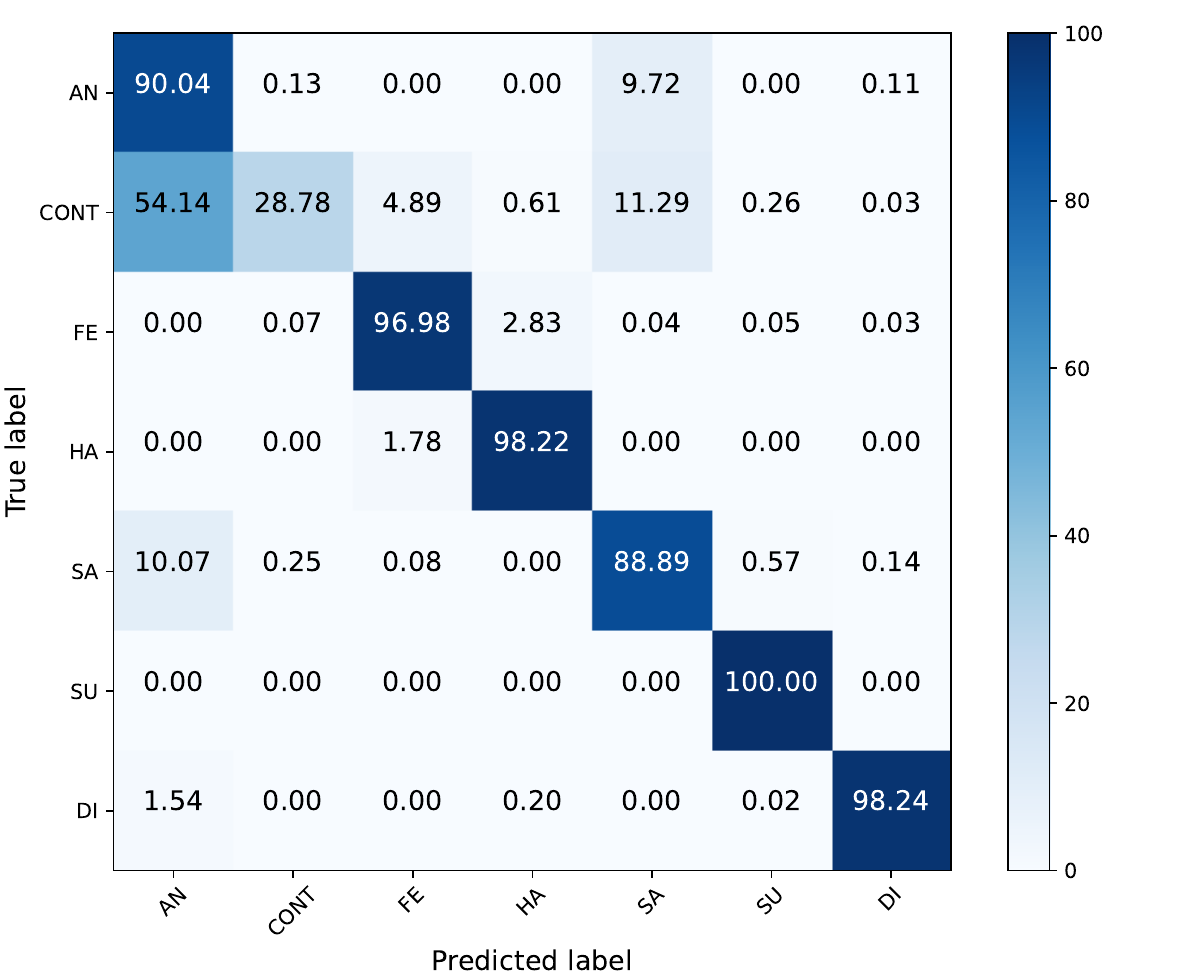}
    \caption{ }
    \label{fig:cfm-ck_t2ft_ftd}
\end{subfigure}
 \caption{Confusion matrices for seven facial expressions by (a) MotionClip and (b) our method on CK+.} 
  \label{fig:cfm-ck}
\end{figure}
\begin{figure}[ht!]
%\centering
%%%%%%%%%% Motion clip %%%%%%%%
\begin{subfigure}{0.48\linewidth}
    \includegraphics[scale=0.258]{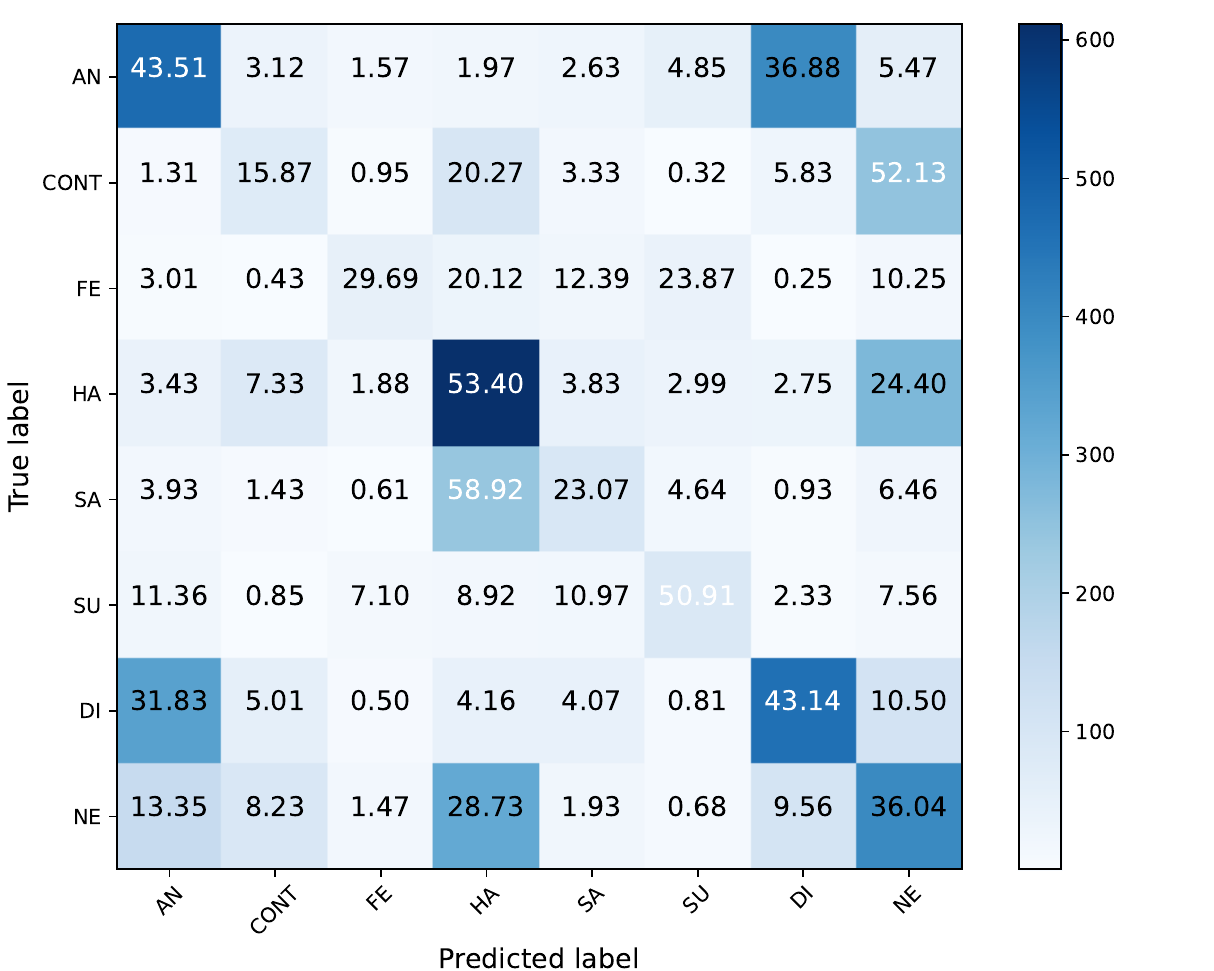}
    \caption{}
    \label{fig:cfm-celebv-hq-motionclip}
\end{subfigure}
%%%%%%%%%% T2FT + FTD %%%%%%%%
\begin{subfigure}{0.45\linewidth}
    \includegraphics[scale=0.25]{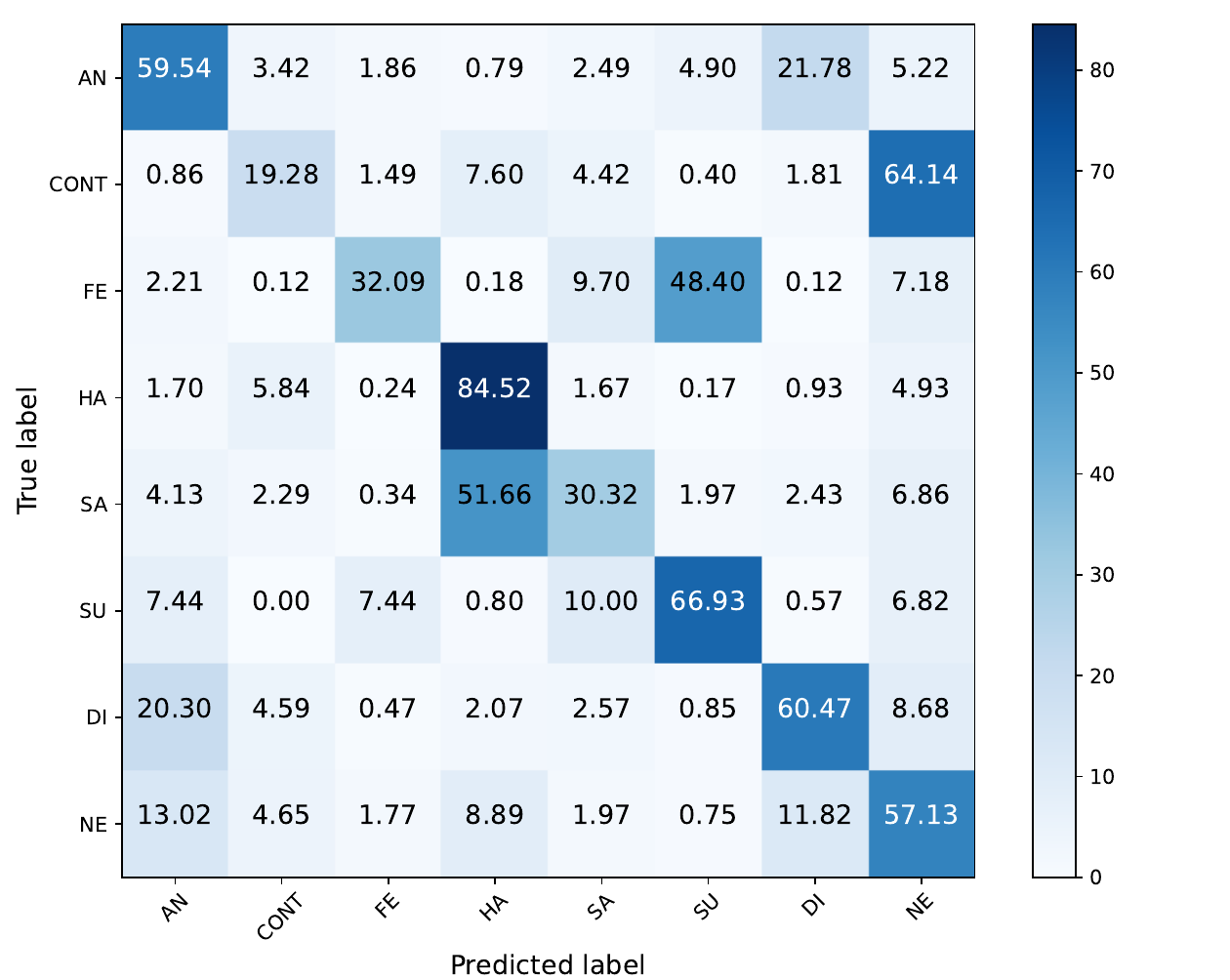}
    \caption{ }
    \label{fig:cfm-celebv-hq_t2ft_ftd}
\end{subfigure}
%\belowcaptionskip = -10pt
 \caption{Confusion matrices for eight facial expressions by (a) MotionClip and (b) our method on CelebV-HQ.}
  \label{fig:cfm-celevhq}
\end{figure}

%%%%%%%%%%%%%%%% Table OF T2FET %%%%%%%%%%
\begin{table}[ht]
    \centering
    %\caption{Comparison of accuracy between our method and MotionClip on CK+ dataset}
    \caption{COMPARISON OF ACCURACY BETWEEN OUR METHOD AND MOTIONCLIP ON THE CK+ DATASET}
    \label{tab:t2fet-ck}
    \begin{tabular}{llll}
    \hline 
    \textbf{Method} & \textbf{Acc}$_1$ (\%) $\uparrow$  & \textbf{Acc}$_2$ (\%) $\uparrow$ & \textbf{Gmean} (\%) $\uparrow$ \\
    \hline 
    Ground-truth & 99.69 $\pm$ 0.0001 & 99.39 $\pm$ 0.0002 & 99.69 $\pm$ 9.3e$^{-5}$ \\ 
    MotionClip & 52.00 $\pm$ 0.0058 & 20.00 $\pm$ 0.0082 & 40.48 $\pm$ 0.0060\\ 
     \textbf{Ours}  & \textbf{91.44 $\pm$ 0.0024} & \textbf{84.03 $\pm$ 0.0034} & \textbf{80.30 $\pm$ 0.0064}\\ 
    \hline 
    \end{tabular}
\end{table}
\vskip -1em
\begin{table}[ht]
    \centering
    %\caption{Comparison of accuracy between our method and MotionClip on CelebV-HQ dataset}
    \caption{COMPARISON OF ACCURACY BETWEEN OUR METHOD AND MOTIONCLIP ON THE CELEBV-HQ DATASET}
    \label{tab:t2fet-celebv}
    \begin{tabular}{llll}
    \hline 
    \textbf{Method} & \textbf{Acc}$_1$ (\%) $\uparrow$ & \textbf{Acc}$_2$ (\%) $\uparrow$ & \textbf{Gmean} (\%) $\uparrow$\\
    \hline 
     Ground-truth &  93.67 $\pm$ 0.0000 & 87.89 $\pm$ 0.0000 & 95.04 $\pm$ 1.1e$^{-16}$\\ 
     MotionClip & 40.00 $\pm$ 0.0039 & 13.73 $\pm$ 0.0056 & 34.42 $\pm$ 0.0069\\
     \textbf{Ours} & \textbf{58.24 $\pm$ 0.0028} & \textbf{33.45 $\pm$ 0.0052} & \textbf{46.47 $\pm$ 0.0065}\\ 
    \hline 
    \end{tabular}
\end{table}

%%%%%%%%%%%%%%%% TABLE OF GENERATED FACIAL EXPRESSION IMAGES %%%%%%%%%%%%%%
\begin{table}[ht]
    \centering
    %\caption{Result obtained when our method is combined with ROME or CVTHead for texture generation on CK+ and CelebV-HQ dataset, respectively. Note that $\downarrow$ indicates that the lower is better, whereas $\uparrow$ denotes that the higher is better.}
    \caption{RESULT OBTAINED BY COMBINING OUR METHOD WITH ROME OR CVTHEAD FOR TEXTURE GENERATION ON THE CK+ AND CELEBV-HQ DATASETS, RESPECTIVELY. NOTE THAT $\downarrow$ INDICATES THAT LOWER IS BETTER, WHEREAS $\uparrow$ DENOTES THAT THE HIGHER IS BETTER.}
    \label{tab:texture-celebv-ck}
    \begin{tabular}{l@{\hspace*{0.12cm}}ll@{\hspace*{0.14cm}}l@{\hspace*{0.14cm}}l@{\hspace*{0.14cm}}l}
    \hline
    \textbf{Dataset} & \textbf{Method} & L$_{1}\downarrow$ & PSNR${\uparrow}$ & LPIPS${\downarrow}$ & MS\_SSIM${\uparrow}$ \\
    \hline
    CK+ & Ours+ROME & 0.199 & 9.939 & 0.490 & 0.347 \\
    & \textbf{Ours+CVTHead} & \textbf{0.005} & \textbf{33.74} & \textbf{0.021} & \textbf{0.978} \\
    \hline 
    CelebV-HQ & Ours+ROME & 0.168 & 11.316 & 0.505 & 0.418 \\
    & \textbf{Ours+CVTHead} & \textbf{0.003} & \textbf{37.08} & \textbf{0.010} & \textbf{0.990} \\
    \hline 
    \end{tabular}
\end{table}

%%%%%%%%%%%%%%%% FIGURE OF GENERATED FACIAL EXPRESSION IMAGES %%%%%%%%%%%%
\begin{figure*}[ht!]
    \centering
    \includegraphics[scale=0.55]{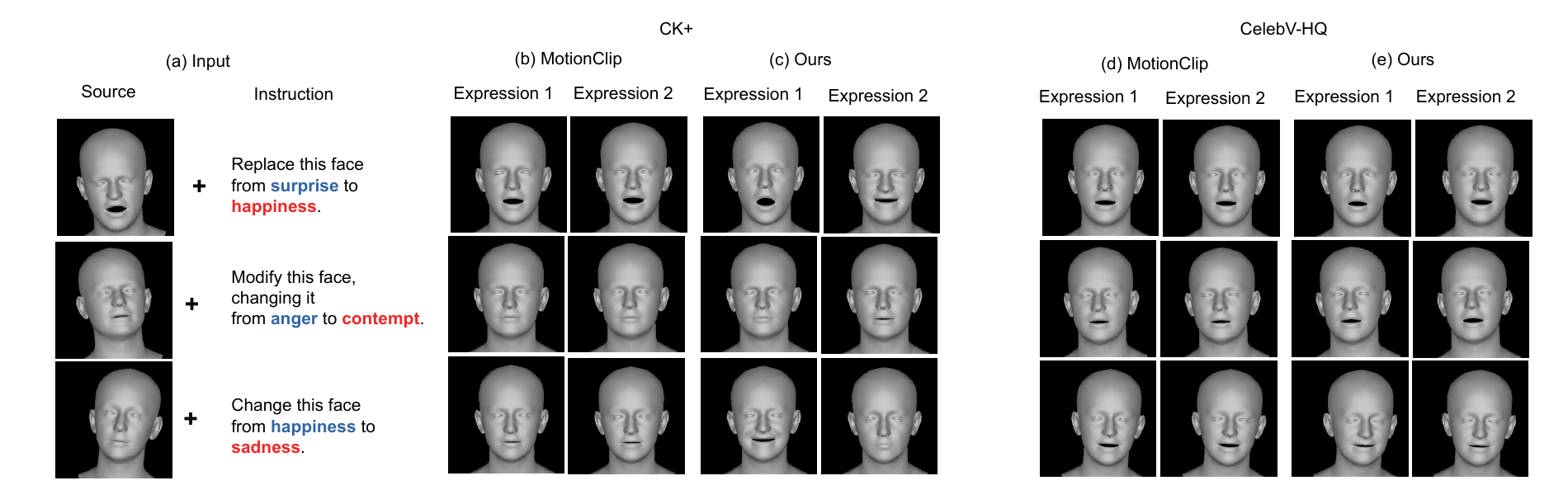}
    \caption{Qualitative comparison between 3D expressive heads that were generated by MotionClip and our method, respectively. The input, consisting of a shape image and instruction, was used for generating expressions 1 and 2 using either MotionClip (b) or Ours (c) on CK+ and CelebV-HQ (d, e), respectively.}
    \label{fig:celebv-ck-shape}
\end{figure*}
\begin{figure*}[ht!]
    \centering
    \includegraphics[scale=0.4]{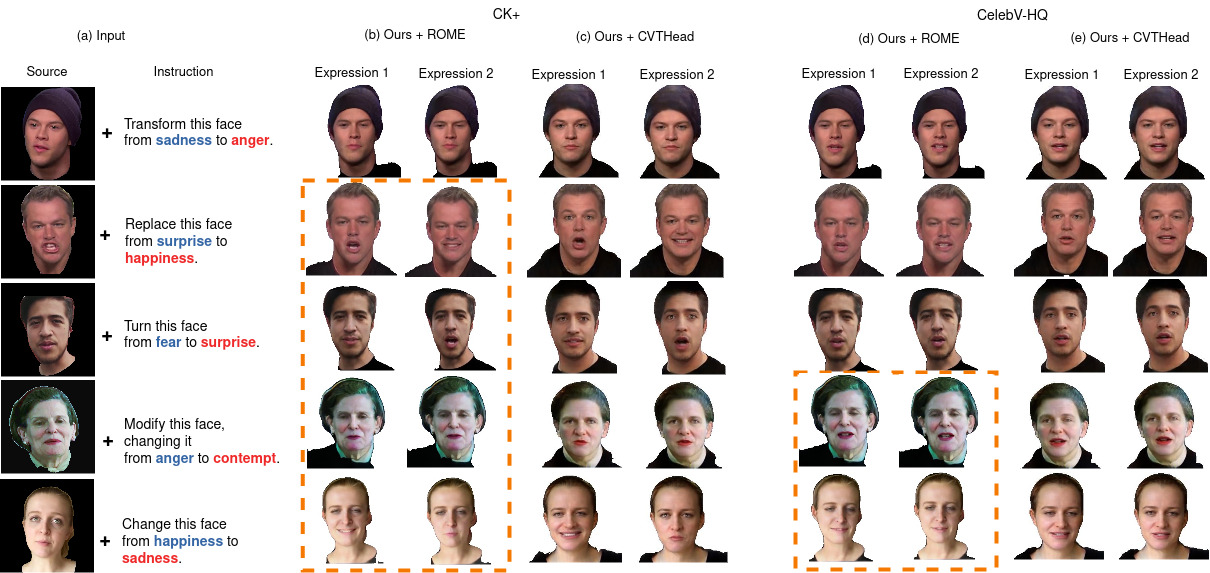}
    \caption{
    Qualitative comparison between generated 3D expressive faces when our method is combined with texture generation \cite{Khakhulin2022ROME, ma2023cvthead} to generate the desired expressions. The input (a), consisting of the source image and instruction, was used for generating expressions 1 and 2 using either Ours+ROME (b) or Ours+CVTHead (c) on CK+ and CelebV-HQ (d, e), respectively.}
    \label{fig:ck-celebv-texture}
\end{figure*}
\begin{figure*}[ht!]
    \centering
    \includegraphics[scale=0.32]{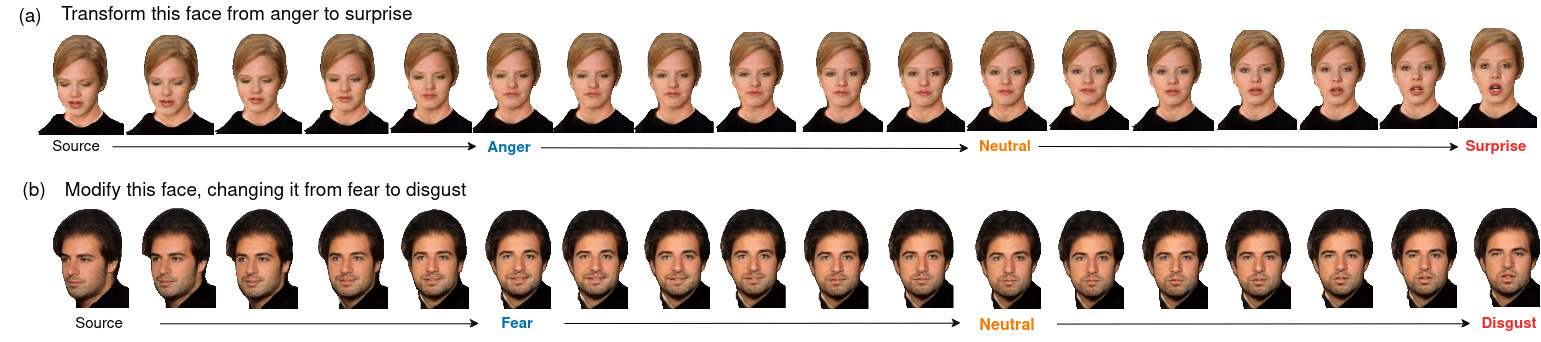}
    \caption{Successive snapshots of the generated 3D expressive faces that include a neutral expression between two facial expressions instructed by text prompts.}
    \label{fig:ck-long-sequence-with-neutral}
\end{figure*}
% %%%%%%%%%%%%%%%%%%%% RESULT %%%%%%%%%%%%%%%%%%%%%%%
% \subsection{Results}
{As the present study aims to generate an avatar with a 3D facial expression from a 2D image of a face and transform the facial emotion according to the given instructions, two major challenges arise (1) facial expression transition based on expression information given by the instruction and (2) rendering the facial appearance such that it corresponds with the desired facial expression. Two experiments were prepared to evaluate the results obtained with the proposed framework both quantitatively and qualitatively.   
\subsubsection{Quantitative Comparison}
Two tasks are prepared for quantitative comparison. 
\\
\hspace*{10pt}\textit{Facial Expression Transition} \\
The DECA encoder extracts the shape, facial expression, and pose from the source image. At the same time, the CLIP encoder combined with IFED and I2FET was used to interpret the text instruction: the initial facial expression, the final facial expression, and pose variation. To compare our approach with MotionClip, the first accuracy metric (Acc$_1$) was used to count the correctly classified samples for all expression labels in the instructions. The other one (Acc$_2$) was used to measure the correctly classified samples corresponding to both expression labels presented in each input instruction. The result suggests that the accuracy of our model is higher than that of MotionClip on both datasets as summarized in Table \ref{tab:t2fet-ck}. For example, our model reaches 91.44 \% accuracy for generating an independent expression and 84.03 \% for generating expression sequences on the CK+ dataset. The Gmean of our model is 80.30\%.     

For the CelebV-HQ dataset, which is the real-world dataset, our model scores 58.24 \%, 33.45 \%, and 46.47 \% for Acc$_1$, Acc$_2$, and Gmean, respectively, as indicated in Table \ref{tab:t2fet-celebv}. This result suggests that our model outperforms MotionClip for learning the facial expression parameters. 

\textit{Visualization with t-SNE}: Fig. \ref{fig:tSNE} shows three learned latent representations, i.e. with MotionClip, I2FET, and I2FET with IFED, for seven facial expressions. These visual representations indicate that segregation between facial expressions is highly distinctive when IFED is combined with I2FET, compared with either MotionClip or I2FET alone.
\\
\hspace*{10pt}
\textit{Confusion Matrices Analysis}: Fig. \ref{fig:cfm-ck} and Fig. \ref{fig:cfm-celevhq} show that IFED enhances the accuracy across most classes compared to MotionClip on both datasets. Yet, a few case of misclassification occurred between certain classes of expressions, such as anger and sadness, or contempt and anger with sadness on CK+. In real-world datasets such as CelebV-HQ, some misclassification occurs between classes, particularly those with challenging expressions such as fear and surprise, or contempt and neutral.

\textit{Face Rendering}: 
For the face-rendering task, FET was combined with state-of-the-art models \cite{Khakhulin2022ROME, ma2023cvthead} for rendering facial appearances that correspond to specific expressions requested by instruction. This was accomplished by integrating whether ROME \cite{Khakhulin2022ROME} or CVTHead \cite{ma2023cvthead} with facial expression transition. In this case, DECA extracts facial parameters to render geometrical information for a specific face. Instead of shapes, the RGB images and instructions are used to evaluate the performance. The L$_1$ loss, Peak Signal to Noise Ratio (PSNR), Learned Perceptual Image Patch Similarity (LPIPS) \cite{zhang2018perceptual}, and Multi-Scale Structured Similarity (MS-SSIM) were used to measure the quality of reconstruction between the target image and synthesized image in the generated expression. As indicated by Table \ref{tab:texture-celebv-ck}, our model with CVTHead produces the best result.

\subsubsection{Qualitative Comparison}
\label{sec:result:quality}
In a similar vein, two tasks were prepared for qualitative comparisons.\\    
\hspace*{10pt} \textit{Facial Expression Transition}:
Fig. \ref{fig:celebv-ck-shape} compares sequences of facial expressions, generated by Motionclip and our method on CK+ and CelebV-HQ, respectively. These results clearly show that our method can generate facial expressions that convey more accurate expressions than MotionClip. 
\\
\hspace*{10pt}\textit{Face Rendering}: 
The resulting expressions can be qualitatively compared by viewing and deciding which expression looks better and more natural than the other. Fig. \ref{fig:ck-celebv-texture} shows the generated facial expressions based on learning the latent space of facial parameters on both datasets. The viewer may notice the difference between (b) and (c) for the CK+ dataset, and between (d) and (e) for the CelebV-HQ dataset, respectively, in terms of facial expressions and facial texture. 
For instance, the facial transition from surprise to happiness seen in the 2nd row is clearer in (c) than in (b), and the facial texture seen in the 4th row is much smoother in (c) than in (b).  A similar observation can be made between (d) and (e), suggesting that Ours+CVTHead is superior to than Ours+ROME. 
For instance, in some cases, those enclosed within dashed boxes in Fig. \ref{fig:ck-celebv-texture}, a subtle distortion arose either when the facial region of the source image was occluded or when the instructions describe expressions with significant pose changes, such as transitioning from surprise to happiness or from fear to surprise. This led to inconsistencies, and missing details in the generated facial regions compared to Ours+CVTHead.
A comparison between the two datasets, such as between (b) and (d) or between (c) and (e), reveals that the visual quality of CK+ is better than that of CelebV-HQ. 
 %In addition, Fig. \ref{fig:ck-long-sequence} shows four demos of facial expression transitions in sequence, starting from a 2D face that has an arbitrary pose. 
 These results confirm that our model has outstanding ability to transform facial expressions, learned on CK+ or CelebV-HQ, to a specific facial image. 
 \\
 %%% User study %%%
 \hspace*{0.3cm}In addition, a user study was carried out by requesting 24 subjects to view 60 randomly paired} images of various identities selected from either CVTHead 
+ MotionClip or CVTHead + Ours. Subjects were asked to choose the expression closest to the expressions specified in the instructions or indicate whether none of them matched the provided expressions. Naturally, they were unaware as to which algorithm was used to the presented images. Four subjects viewed each video and voted for the one that more closely resembled the expression in the instructions. In cases in which the votes were tied, an additional subject was asked to make the final decision. Based on these outcomes, CVTHead + Ours achieved 73\% accuracy, whereas CVTHead + MotionClip yielded 58\% accuracy for CK+. For CelebV-HQ, CVTHead + Ours and CVTHead + MotionClip achieved 77\% and 70.5\% accuracy, respectively.

\subsubsection{Inference Time}
During inference, CVTHead + MotionClip takes 3.97 seconds to generate a video, whereas CVTHead + Ours only takes 3.92 seconds on a single NVIDIA RTX A6000, suggesting that our model is a slightly faster than MotionClip.

\subsection{Ablation Study}
\label{sec:ablstu}
This section presents a series of ablation studies that were carried out to evaluate the performance of our model.
\subsubsection{Effectiveness of the Proposed Components}
To demonstrate the effectiveness of each proposed component, we conducted a comparative experiment, as shown in Table \ref{tab:t2fet-ck-celebv-component}. 
It can be observed that the IFED module significantly improves the performance of I2FET compared to I2FET without IFED on both CK+ and CelebV-HQ datasets. Additionally, integration of the vertex loss function with IFED increases Acc$_1$ and Gmean on both datasets. Despite a slight decrease in Acc$_2$ on CK+, it still performs well on CelebV-HQ. 
\begin{table}[ht]
    \centering
    %\caption{Comparison of accuracy between the proposed different components for our model (T2FET) on two datasets}
    \caption{COMPARISON OF THE ACCURACY BETWEEN THE PROPOSED DIFFERENT COMPONENTS FOR OUR METHOD (I2FET) ON THE TWO DATASETS}
    \label{tab:t2fet-ck-celebv-component}
    \begin{tabular}{l@{\hspace*{0.12cm}}l@{\hspace*{0.12cm}}l@{\hspace*{0.12cm}}l@{\hspace*{0.12cm}}l}
    \hline 
    \textbf{Dataset} & \textbf{Method} & \textbf{Acc}$_1$ (\%) $\uparrow$  & \textbf{Acc}$_2$ (\%) $\uparrow$ & \textbf{Gmean} (\%) $\uparrow$ \\
    \hline 
    & w/o IFED  & 76.89 $\pm$ 0.0029 & 63.23 $\pm$ 0.0047 & 63.59 $\pm$ 0.0059\\ 
    & w/ IFED  & 91.30 $\pm$ 0.0017 & \textbf{84.06 $\pm$ 0.0026} & 79.41 $\pm$ 0.0061\\ 
   CK+  &\textbf{w/ IFED}  & \textbf{91.44 $\pm$ 0.0024} & 84.03 $\pm$ 0.0034 & \textbf{80.30 $\pm$ 0.0064}\\ 
   &+\textbf{w/ $\mathcal{L}_{v}$} & & & \\
    \hline 
     & w/o IFED  & 48.35 $\pm$ 0.0037 & 22.1 $\pm$ 0.0040 & 32.88 $\pm$ 0.0058\\
     & w/ IFED  & 57.80 $\pm$ 0.0020 & 32.60 $\pm$ 0.0037 & 46.04 $\pm$ 0.0062\\ 
   CelebV-HQ  & \textbf{w/ IFED} & \textbf{58.24 $\pm$ 0.0028} & \textbf{33.45 $\pm$ 0.0052} & \textbf{46.47 $\pm$ 0.0065}\\ 
   &+\textbf{w/ $\mathcal{L}_{v}$} & & & \\
    \hline 
    \end{tabular}
\end{table}

\subsubsection{Impact of Three Loss Functions}
\begin{figure}[ht!]
%\centering
\begin{subfigure}{0.48\linewidth}
    \includegraphics[scale=0.258]{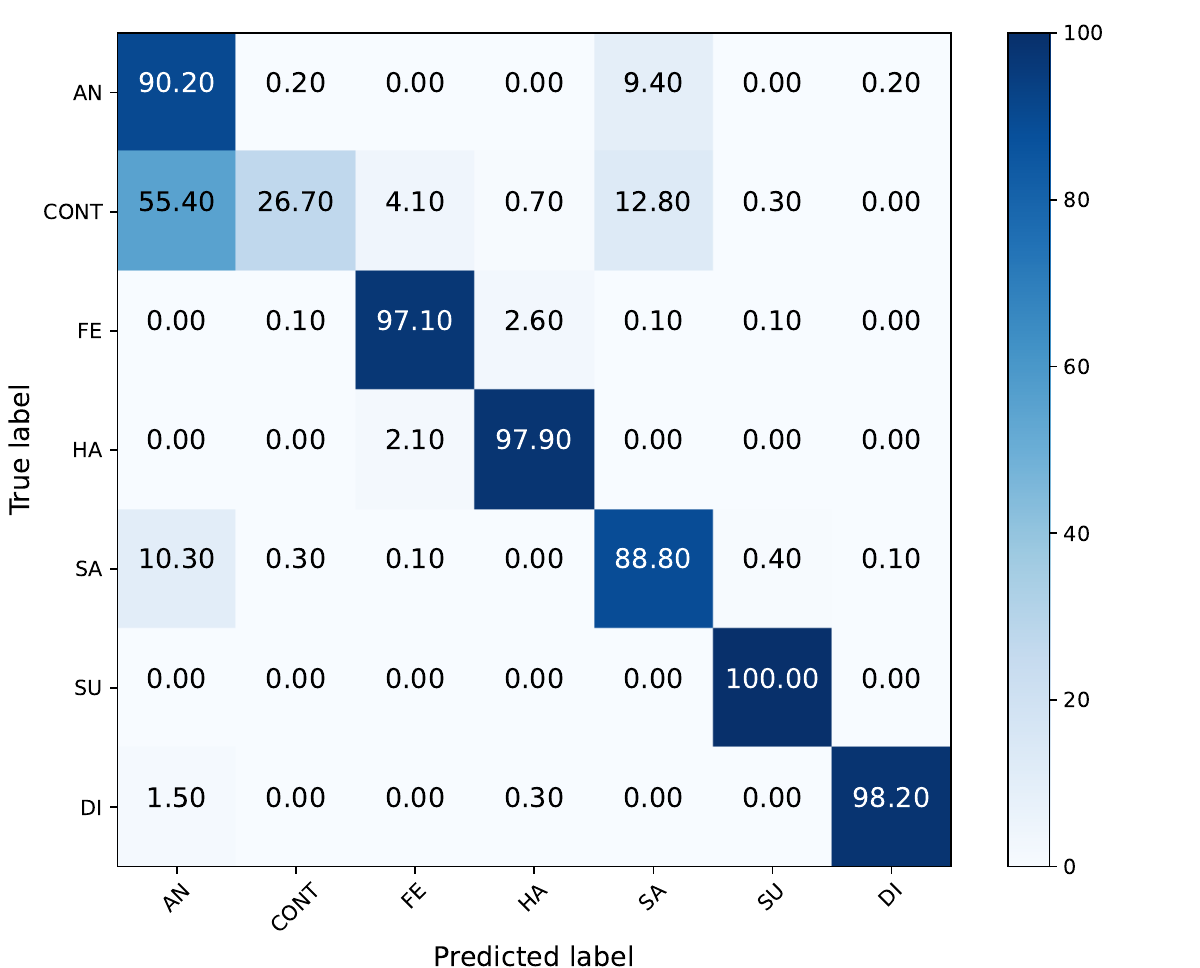}
    \caption{ }
    \label{fig:cfm-ck-non-vert}
\end{subfigure}
%%%%%%%%%% T2FT + TFED %%%%%%%%
\begin{subfigure}{0.45\linewidth}
    \includegraphics[scale=0.254]{figs/ck_t2ft_ftd.pdf}
    \caption{ }
    \label{fig:cfm-ck_vert}
\end{subfigure}
 \caption{Confusion matrices for seven facial expressions by I2FET (a) without $\mathcal{L}_{v}$ and (b) with $\mathcal{L}_{v}$ on CK+.} 
  \label{fig:cfm-ck_vert-non-vert}
\end{figure}

\begin{figure}[ht!]
%\centering
\begin{subfigure}{0.48\linewidth}
    \includegraphics[scale=0.258]{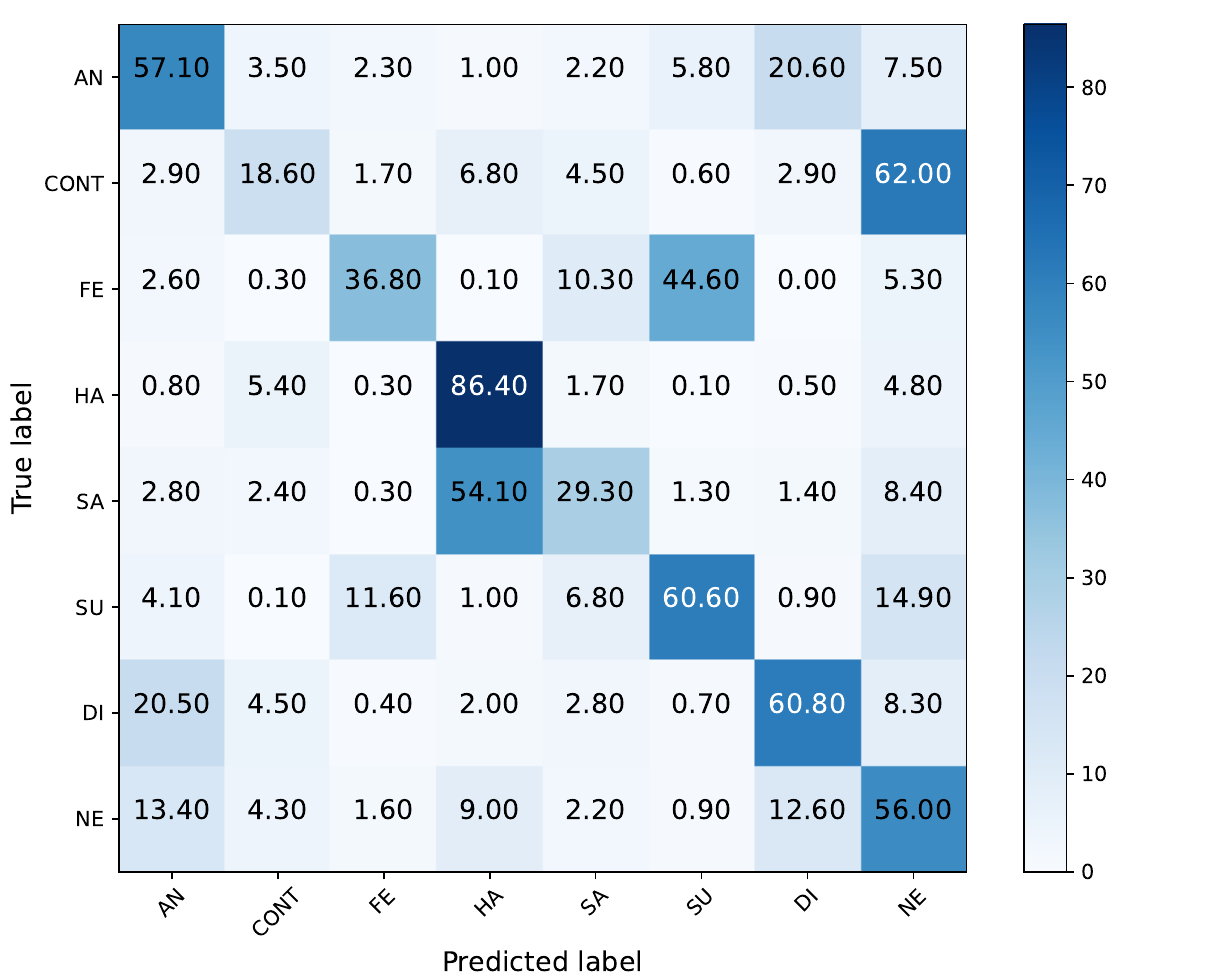}
    \caption{}
    \label{fig:cfm-celebv-hq-non-vert}
\end{subfigure}
%%%%%%%%%% T2FT + FTD %%%%%%%%
\begin{subfigure}{0.45\linewidth}
    \includegraphics[scale=0.25]{figs/celebv_t2ft_ftd.pdf}
    \caption{ }
    \label{fig:cfm-celebv-hq_vert}
\end{subfigure}
%\belowcaptionskip = -10pt
 \caption{Confusion matrices for eight facial expressions by I2FET (a) without $\mathcal{L}_{v}$ and (b) with $\mathcal{L}_{v}$ on CelebV-HQ.}
  \label{fig:cfm-cfm-celebv-hq_vert-non-vert}
\end{figure}

To analyze the impact of each loss function in I2FET, an experiment was carried out to compare the performance of the model in various scenarios. Specifically, IFED was integrated with the I2FET model and each loss function was incrementally added to determine its impact on the model performance of the model. The results in Table \ref{tab:t2fet-ck-celebv-loss} indicate that the performance improved when the pose loss function integrated with the expression loss function, whereas the vertex loss function yielded a slight improvement when used alongside both the pose and expression loss functions.
In this study, the vertex loss was integrated with other loss functions to capture the geometry of the face, to enhance the accuracy of facial expression recognition. Fig \ref{fig:cfm-ck_vert-non-vert} and \ref{fig:cfm-cfm-celebv-hq_vert-non-vert} are the confusion matrix of our model when the total loss function integrated with $\mathcal{L}_v$ or without $\mathcal{L}_v$ on the CK+ and CelebV-HQ datasets. Note that the fear accuracies for both datasets rather decrease slightly when the vertex loss is applied. It is known that the fear expression often overlaps with other expressions (e.g., surprise and happiness), thereby making it challenging to distinguish. Moreover, since the fear portion of CK+ is much larger than that of CelebV-HQ in Fig. 5, it appears that confusion during the facial expression transition has a reverse impact on Acc$_2$ as shown in Table \ref{tab:t2fet-ck-celebv-loss}  

%%% 
%%%%%%%%%%%%%%%%%%%%%%%%%%%%%%%%%%%%%%%%%%%%%%

\begin{table}[ht]
    \centering
   
    \caption{COMPARISON OF THE IMPACT OF LOSS FUNCTIONS ON THE PERFORMANCE OF OUR METHOD (I2FET) ACROSS THE TWO DATASETS}
    \label{tab:t2fet-ck-celebv-loss}
    \begin{tabular}{l@{\hspace*{0.12cm}}l@{\hspace*{0.14cm}}l@{\hspace*{0.12cm}}l@{\hspace*{0.12cm}}l}
    \hline 
    \textbf{Dataset} & \textbf{Method} & \textbf{Acc}$_1$ (\%) $\uparrow$  & \textbf{Acc}$_2$ (\%) $\uparrow$ & \textbf{Gmean} (\%) $\uparrow$ \\
    \hline 
    & \textbf{$\mathcal{L}_{e}$} & 66.80 $\pm$ 0.0056 & 45.96 $\pm$ 0.0112 & 52.42 $\pm$ 0.0730\\ 
    %& & & & \\
    & \textbf{$\mathcal{L}_{e}$}+$\mathcal{L}_{p}$  & 91.30 $\pm$ 0.0017 & \textbf{84.06 $\pm$ 0.0026} & 79.41 $\pm$ 0.0061\\ 
    %& \color{red}+ $\mathcal{L}_{pose}$ & & & \\
    %& & & & \\
   CK+  & \textbf{$\mathcal{L}_{e}$}+$\mathcal{L}_{p}$+$\mathcal{L}_{v}$  & \textbf{91.44 $\pm$ 0.0024} & 84.03 $\pm$ 0.0034 & \textbf{80.30 $\pm$ 0.0064}\\ 
    \hline 
     & \textbf{$\mathcal{L}_{e}$}  & 37.57 $\pm$ 0.0052 & 11.99 $\pm$ 0.0326 & 25.82 $\pm$ 0.0114\\
     
     & \textbf{$\mathcal{L}_{e}$}+$\mathcal{L}_{p}$& 57.80 $\pm$ 0.0020 & 32.60 $\pm$ 0.0037 & 46.04 $\pm$ 0.0062\\

   CelebV-HQ  & \textbf{$\mathcal{L}_{e}$}+$\mathcal{L}_{p}$+$\mathcal{L}_{v}$ & \textbf{58.24 $\pm$ 0.0028} & \textbf{33.45 $\pm$ 0.0052} & \textbf{46.47 $\pm$ 0.0065}\\
    \hline 
    \end{tabular}
\end{table}

%%%%%%%%%%%%%%%%%%%%%%%%%%%%%%%%%%%%%%%%%%%%%%%%%%%%%%%%%%%
\subsubsection{Impact of IFED on the Training Process} 
\begin{figure}[ht!]
\belowcaptionskip = -7pt
\centering
%%%%%%%%%% CK+ %%%%%%%%
\begin{subfigure}{0.5\linewidth}
    \includegraphics[scale=0.29]{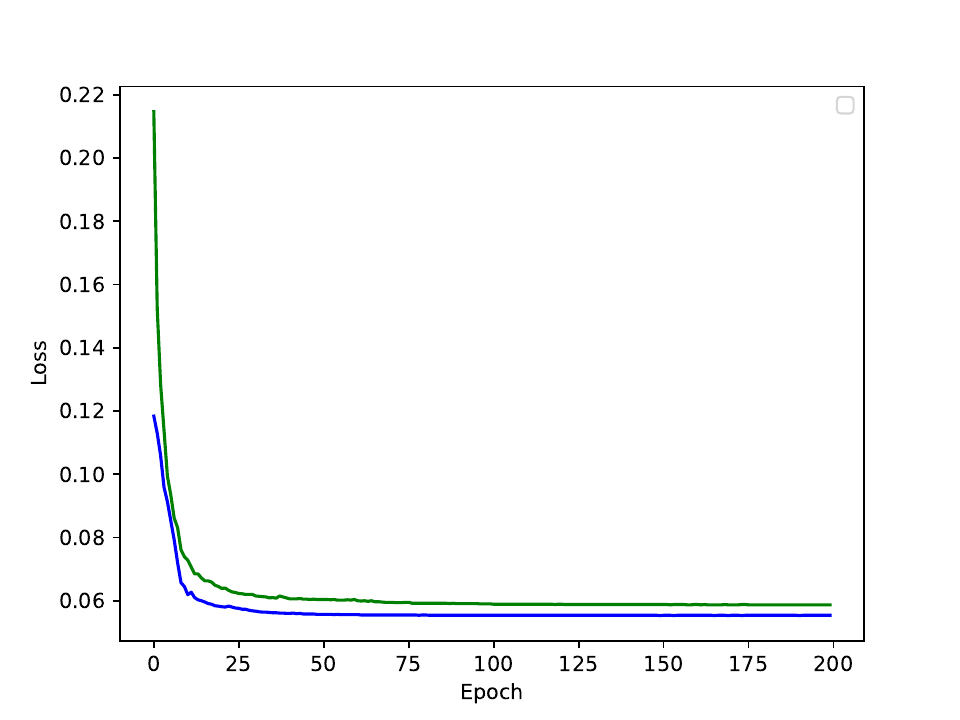}
    \caption{ }
    \label{fig:tfeg-ck-loss}
\end{subfigure}
%%%%%%%%%% CelebV-HQ %%%%%%%%
\begin{subfigure}{0.45\linewidth}
    \includegraphics[scale=0.29]{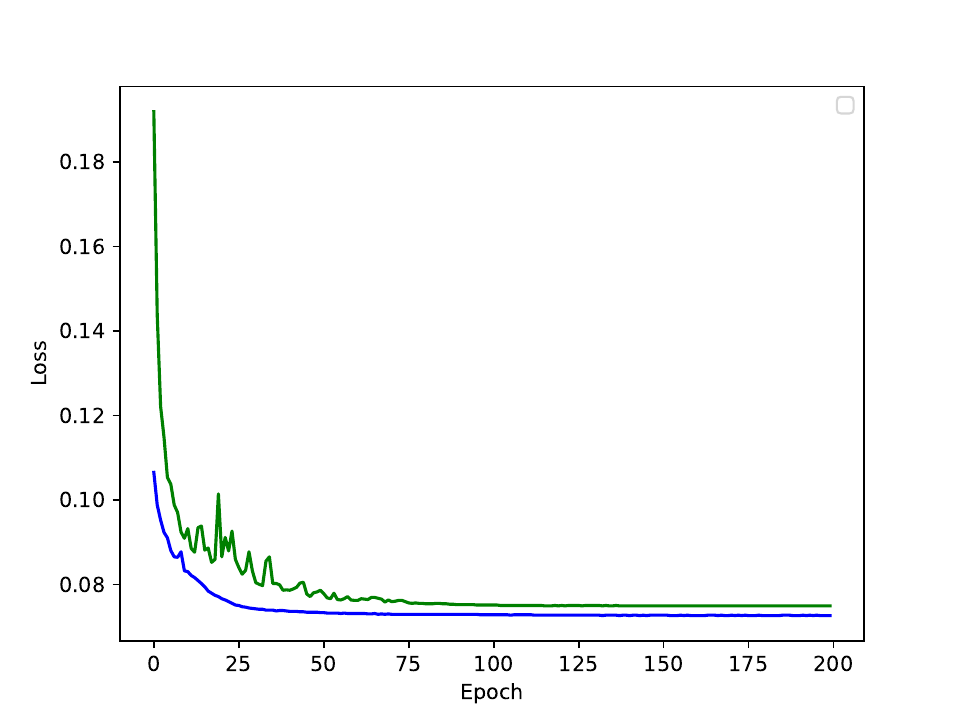}
    \caption{ }
    \label{fig:tfeg-celebv-hq-loss}
\end{subfigure}
\caption{Comparison of the validation loss during the training process of I2FET w/ and w/o IFED, respectively, on (a) CK+ and (b) CelebV-HQ. The green curves represent I2FET without IFED, whereas the blue curves are the results for I2FET with IFED.}
\label{fig:loss}
\end{figure}

This experiment was conducted to compare the two loss functions with and without incorporating the IFED module in I2FET. Fig. \ref{fig:tfeg-ck-loss} and Fig. \ref{fig:tfeg-celebv-hq-loss} show the results for CK+ and CelebV-HQ, respectively, confirming that I2FED plays an important role.
\subsubsection{Effect of IFED on Intensity of Facial Expression}
\begin{figure}[ht!]
\centering
\includegraphics[scale=0.5]{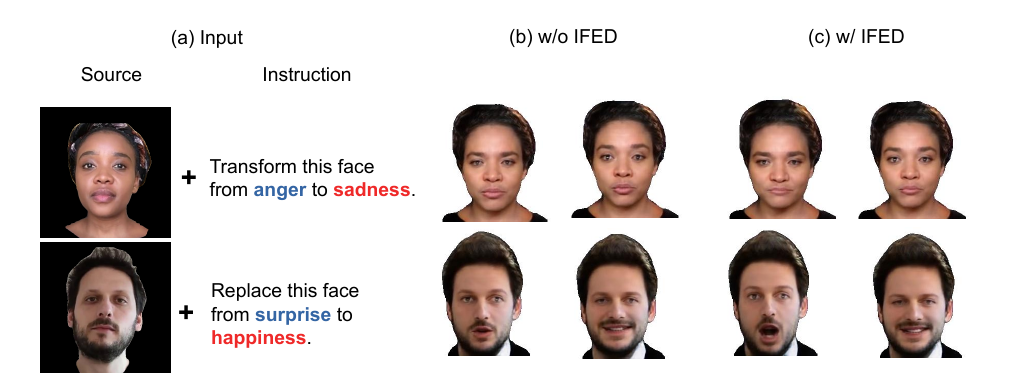}
    \caption{Qualitative comparison between 3D expressive faces generated (a) w/o and (b) w/ IFED.}
    \label{fig:abs_w_and_wo}
\end{figure}

Fig. \ref{fig:abs_w_and_wo} demonstrates that IFED helps I2FED to generate high-intensity facial expressions.

%%%%% Loss %%%%%%%
%%%%% Intensity Expression %%%%

%%%%%%%%%
\subsubsection{Impact of Quantity of CAFT Layer} 
The experiments were conducted to evaluate the results of the proposed I2FET model by designing the IFED module with varying numbers of CAFT layers. We fixed the number of transformer layers to 1 in both branches and adjusted the number of CAFT layers. The results in Table \ref{tab:nblock-t2fet-ck-celebv} indicate that increasing the quantity of CAFT layers does not improve the performance of the proposed I2FET.

\subsubsection{Impact of Quantity of Transformer Encoder in Facial Parameter Branch and Text Branch} 
In this experiment, we utilized different configurations by adjusting the number of transformer encoder layers of IFED. 
The findings suggest that the I2FET model performs better when \textbf{N} = 2 and \textbf{M} = 1 compared to other configurations, particularly in terms of Acc$_1$ and G-mean, as shown in Table \ref{tab:ntransformer-t2fet}. 

\begin{table}[ht]
    \centering
    %\caption{Comparison of accuracy of the T2FET method with varying numbers of CAFT layers in TFED on both datasets}
    \caption{COMPARISON OF THE ACCURACY OF THE I2FET METHOD ON BOTH DATASETS BY VARYING THE NUMBER OF CAFT LAYERS IN I2FED}
    \label{tab:nblock-t2fet-ck-celebv}
    \begin{tabular}{l@{\hspace*{0.12cm}}c@{\hspace*{0.12cm}}l@{\hspace*{0.12cm}}l@{\hspace*{0.12cm}}l}
    \hline 
    \textbf{Dataset} & \textbf{\#Layers} & \textbf{Acc}$_1$ (\%) $\uparrow$  & \textbf{Acc}$_2$ (\%) $\uparrow$ & \textbf{Gmean} (\%) $\uparrow$ \\
    \hline 
      & 1 & \textbf{91.36 $\pm$ 0.0023} & \textbf{84.13 $\pm$ 0.0046} & \textbf{79.65 $\pm$ 0.0080}\\ 
    %\hline 
  CK+  & 2 & 88.95 $\pm$ 0.0019 & 80.95 $\pm$ 0.0038 & 78.72 $\pm$ 0.0071\\ 
    \hline 
      & 1 & \textbf{58.25 $\pm$ 0.0034} & \textbf{33.03 $\pm$ 0.0056} & \textbf{46.46 $\pm$ 0.0066}\\
   CelebV-HQ & 2 & 55.68 $\pm$ 0.0034 & 29.82 $\pm$ 0.0066 & 42.59 $\pm$ 0.0085\\ 
    \hline
    \end{tabular}
\end{table}
\vskip -1em
\begin{table}[ht]
    \centering
    %\caption{Comparison of accuracy of the proposed T2FET method with the numbers of Transformer layer for facial parameters branch (\textbf{N}) and text branch (\textbf{M}) in TFED on CK+}
    \caption{COMPARISON OF THE ACCURACY OF THE PROPOSED METHOD WITH THE NUMBER OF TRANSFORMER LAYERS FOR FACIAL PARAMETERS BRANCH (\textbf{N}) AND TEXT BRANCH (\textbf{M}) IN TFED ON CK+}
    \label{tab:ntransformer-t2fet}
    \begin{tabular}{l@{\hspace*{0.12cm}}l@{\hspace*{0.12cm}}llll}
    \hline 
    Mode & \textbf{N} & \textbf{M} & \textbf{Acc}$_1$ (\%) $\uparrow$ & \textbf{Acc}$_2$ (\%) $\uparrow$ & \textbf{Gmean} (\%) $\uparrow$\\
    \hline  
     A & 1 & 1 & 91.36 $\pm$ 0.0023 & \textbf{84.13 $\pm$ 0.0046} & 79.67 $\pm$ 0.008\\ 
   % \hline
     B & 2 & 1 & \textbf{91.44 $\pm$ 0.0024} & 84.03 $\pm$ 0.0034 & \textbf{80.30 $\pm$ 0.0064}\\
    % \hline
     C & 1 & 2 & 91.27 $\pm$ 0.0019 & 83.84 $\pm$ 0.0037 & 79.69 $\pm$ 0.0052\\
    % \hline
     D & 2 & 2 & 89.07 $\pm$ 0.0021 & 81.13 $\pm$ 0.0034 & 78.71 $\pm$ 0.0070\\
    \hline
    \end{tabular}
\end{table}

\subsection{Longer Facial Expression Generation with Neutral Expressions}

According to appraisal theory \cite{Scherer}, which is one of the major emotion theories, the dynamic of emotion is described as the sequential check theory of emotion differentiation. In other words, the differentiation of emotional states of humans is the result of a sequence of specific stimulus evaluation appraisal checks, rather than an action of hopping from one emotional state to another in a discrete manner. In our framework, the transition of the facial expression from A to B is similar to the differentiation of emotional states in a sequence: starting from A facial expression with strong emotion a neutral expression and then moving to strong facial expression B. In addition to appraisal theory, the present approach is suitable for investigating facial behaviors within the pleasure-arousal (P-A) space \cite{russel1997}, where the neutral expression serves as a transitional expression between two specific expressions. For this purpose, a neutral encoder-decoder (NED) was adopted to generate neutral faces. The aim thereof was to improve the performance of our framework when a neutral expression is not available in the instruction to increase its flexibility for extending facial expression sequences. 
%%% Neutral dataset 
\subsubsection{Implementation Details}
In this scenario, two datasets containing the parameters of neutral expressions and poses were constructed the pretrained DECA to extract the data from the CelebV-HQ and CK+ datasets. As the CK+ dataset does not contain neutral expressions, the first frame of each video was extracted and considered as the representative neutral case. Afterward, NED was trained on these two neutral datasets. 
\subsubsection{Network Architecture}
The neutral encoder-decoder model was constructed with fully connected layers. The encoder consisted of four fully connected layers, whereas the decoder was structured with four fully connected layers dedicated to pose and expression, respectively.  
\subsubsection{Analysis}
Similarly, we also conducted experiments to evaluate neutral expression generation for both quantitative and qualitative comparisons.
\\
%%% Quantitative comparison %%%
\hspace*{13pt}\textit{Quantitative comparison}:
Specifically,  the accuracy of NED on the CK+ and CelebV-HQ datasets was 49.5\% and 51.33\% for generating the neutral expression, respectively. These results are presented in 
 Table \ref{tab:texture-neutral}, which confirms that NED + CVTHead outperforms NED + ROME. 
 
\begin{figure}[ht!]
    \centering
    %\belowcaptionskip = -10pt
    \includegraphics[scale=0.5]{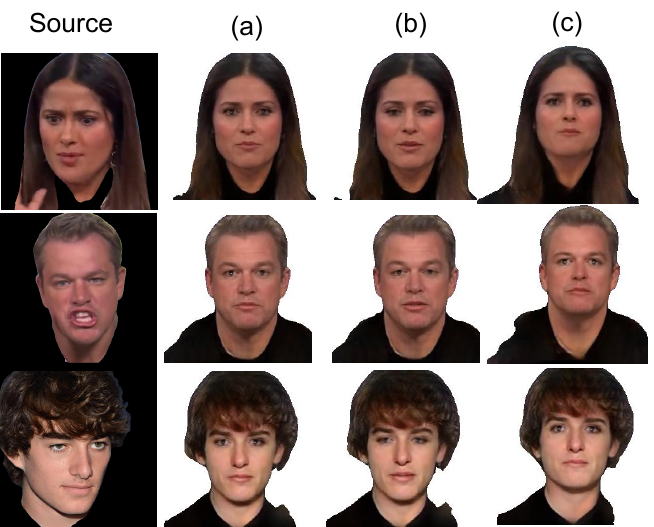}
    \caption{Samples of facial images with a neutral expression generated by NED+CVTHead. The input images are shown in the leftmost column (Source). The images generated using CK+ dataset (controllable environment) are shown in (a), whereas those using CelebV-HQ dataset (non-controllable environment) are shown in (b). On the other hand, the generated 3D face with FLAME is depicted by setting the expression parameter to zero in (c). Note that subtle differences exist in the image quality between the three generated cases.}
    \label{fig:ck-celebv-ned}
\end{figure}

%%% Qualitative comparison %%%
\hspace*{5pt}\textit{Qualitative comparison}:
Fig. \ref{fig:ck-celebv-ned} visualizes neutral expressions generated by NED + CVTHead on CK+ and CelebV-HQ. Moreover, we also illustrate neutral expressions generated by FLAME with the expression parameter set to zero, as shown in Fig. \ref{fig:ck-celebv-ned}c.  

The results show that the facial units differ, depending on whether the environment is controllable or non-controllable, or whether FLAME was used. This indicates that neutral expressions are more complex in real-world scenarios, leading us to prefer using a neural encoder-decoder instead of the expression parameter settings provided by FLAME

%%%%%%%%%%%%%% NEUTRAL EXPRESSION %%%%%%%%%%%%%%%%%
\begin{table}[ht!]
    \centering
   % \caption{Result obtained when NED is combined with ROME or CVTHead for texture generation on CK+ and CelebV-HQ, respectively.}
     \caption{RESULT OBTAINED WHEN NED IS COMBINED WITH ROME OR CVTHEAD FOR TEXTURE GENERATION ON CK+ AND CELEBV-HQ, RESPECTIVELY.}
    \label{tab:texture-neutral}
    \begin{tabular}{l@{\hspace*{0.12cm}}l@{\hspace*{0.12cm}}l@{\hspace*{0.12cm}}l@{\hspace*{0.12cm}}l@{\hspace*{0.12cm}}l}
    \hline
    \textbf{Dataset} & \textbf{Method} & L$_{1}\downarrow$ & PSNR${\uparrow}$ & LPIPS${\downarrow}$ & MS\_SSIM${\uparrow}$ \\
    \hline
    CK+ & NED + ROME & 0.193 & 10.207 & 0.516 & 0.348 \\
    & \textbf{NED + CVTHead} & \textbf{0.004} & \textbf{35.158} & \textbf{0.014} & \textbf{0.987} \\
    \hline
     CelebV-HQ & NED + ROME & 0.183 & 10.818 & 0.498 & 0.407 \\
    & \textbf{NED + CVTHead} & \textbf{0.004} & \textbf{36.347} & \textbf{0.0117} & \textbf{0.989} \\
    \hline  
    \end{tabular}
\end{table}

% \begin{figure*}[ht!]
%     \centering
%     \includegraphics[scale=0.41]{figs/test-FET.pdf}
%     \caption{Example results for longer sequence generation with the insertion of a neutral expression between two specific expressions in given instruction.}
%     \label{fig:ck-long-sequence-with-neutral}
% \end{figure*}

In a user study similar to that described in Section \ref{sec:result:quality}, we asked subjects to determine whether the generated images display neutral expressions. Most subjects succeeded in identifying the neutral expressions produced by our model on CK+, and they had difficulties with only 4\% of the samples generated by the learned model on CelebV-HQ.
%%%%%%%%%%%%%% FACIAL EXPRESSION CLASSIFICATION %%%%%%%%%%%%%%%%%
%%%%%%%%%%%%%%%%%%% FER: TABLE %%%%%%%%%%%%%%%%%%%%%%%%%%
\begin{table*}[ht!]
    \centering
    %\caption{Facial expression classifiers are compared on CK+ and CelebV-HQ with rotated samples at various angles $\{ 0^o, \pm 10^o, \pm 15^o, \pm30^o\}$}
    \caption{COMPARISON OF FACIAL EXPRESSION CLASSIFIERS ON CK+ AND CELEBV-HQ WITH SAMPLES ROTATED AT VARIOUS ANGLES $\{ 0^o, \pm 10^o, \pm 15^o, \pm30^o\}$}
     \label{tab:ck-rotate}
    \begin{tabular}{cccccccccc}
    %\toprule
    \hline
    \multicolumn{1}{c}{} & \multicolumn{1}{c}{} & \multicolumn{2}{c}{\textbf{0$^o$}} & \multicolumn{2}{c}{$\pm$\textbf{10$^o$}}  & \multicolumn{2}{c}{$\pm$\textbf{15$^o$}}  & \multicolumn{2}{c}{$\pm$\textbf{30$^o$}}  \\
    %\hline
    %\midrule
    \cmidrule(rl){3-4} \cmidrule(rl){5-6} \cmidrule(rl){7-8} \cmidrule(rl){9-10}
    \textbf{Dataset} & \textbf{Method} & \textbf{Acc}$_1$ $\uparrow$ & \textbf{G-mean} $\uparrow$ & \textbf{Acc}$_1$ $\uparrow$& \textbf{G-mean} $\uparrow$& \textbf{Acc}$_1$ $\uparrow$ & \textbf{G-mean} $\uparrow$& \textbf{Acc}$_1$ $\uparrow$& \textbf{G-mean} $\uparrow$ \\
    %\midrule 
    \hline
    & MEK \cite{zhang2023leave} & \textbf{100} & \textbf{100} & 86.41 & \textbf{76.20} & 76.01 & 52.16 & 32.56 & 0.0 \\
    %\hline
   CK+  & ResNet-101 \cite{he2016residual} & 99.56 & 99.4 & 84.40 & 73.28 & 79.50 & \textbf{67.6} & 75.29 & \textbf{33.39} \\
    %\hline
    & ResNet-101 with RFL & 99.69 & 99.69 & \textbf{87.18} & 75.85 & \textbf{83.66} & 66.13 & \textbf{77.48} & 31.5\\
    %& with RFL &  & & & & & & & \\
    \hline
    %%% CelebV-HQ
    & MEK \cite{zhang2023leave} & \textbf{99.75} & \textbf{99.83} & \textbf{60.49} & \textbf{55.07} & 49.08& 33.84& 32.05& 9.64\\
    %\hline
   CelebV-HQ &  ResNet-101 \cite{he2016residual} & 37 & 12.77 & 35.77& 13.39& 34.97& 0.0& 28.92 & 0.0 \\
    %\hline
   & ResNet-101  with RFL & 93.67 & 95.04 & 58.54& 49.92& \textbf{51.53} & \textbf{41.22} & \textbf{40.71} & \textbf{26.80}\\
    %& with RFL &  & & & & & & & \\
     \hline
    \end{tabular}
\end{table*}

\subsection{Facial Expression Classification }
\vskip -0.45em
\subsubsection{Implementation}
For ResNet-101 \cite{he2016residual} and ResNet-101 with RFL \cite{Cui2019ClassBalancedLB}, the learning rate was $1e^{-4}$ and the batch size was 64. Adam was used to optimize the models with 50 epochs per cycle. For the MEK model, the configuration in \cite{zhang2023leave} was used to optimize the training process for both the CK+ and CelebV-HQ datasets.
\subsubsection{Analysis}
Based on the results in Table \ref{tab:ck-rotate}, ResNet-101 with RFL was more stable compared to ResNet-101 or MEK in terms of their classification performance, suggesting that ResNet-101 with RFL can handle both imbalanced datasets and rotational transformations fairly well. In particular, MEK achieved the best result on the original images although its performance was affected by rotated images. ResNet-101 with RFL demonstrated more stable performance than ResNet-101 and MEK on CK+. For real-world scenarios such as those in the CelebV-HQ dataset, we obtained similar results. These findings indicate that the performance of facial expression classifiers deteriorates for faces that are rotated by $\pm 15$ degrees or more. In addition, MEK performs better for the original datasets, but worse for the augmented datasets. The performance of ResNet-101 is high on datasets with low imbalance rates, whereas it is low on datasets with high imbalance rates.   
Meanwhile, ResNet-101 with RFL delivers stable performance on imbalanced and rotated datasets. Based on this result, we adopted ResNet-101 by default during evaluation.   

%%% Word collection %%%
% \begin{figure}[ht!]
% \centering
% \includegraphics[scale=0.48]{figs/vocab.pdf}
%     \caption{Example of the generated images by ours in terms of changing the template sentences}
%     \label{fig:vocab}
% \end{figure}
% %%% Failure cases %%%
% \begin{figure}[ht!]
% \centering
% \includegraphics[scale=0.48]{figs/failure_cases.pdf}
%     \caption{Failure results of our method.}
%     \label{fig:failure_cases}
% \end{figure}
\begin{figure}[ht!]
\centering
%%%%%%%%%% CK+ %%%%%%%%
\begin{subfigure}{0.74\linewidth}
    \includegraphics[scale=0.35]{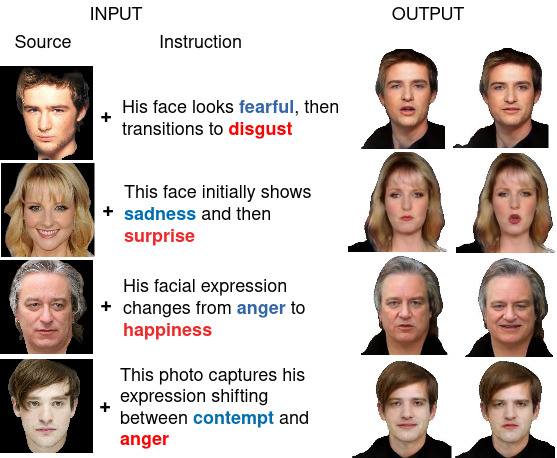}
    \caption{ }
    \label{fig:vocab}
\end{subfigure}
%%%%%%%%%% CelebV-HQ %%%%%%%%
\begin{subfigure}[b]{0.82\linewidth}
    \includegraphics[scale=0.5]{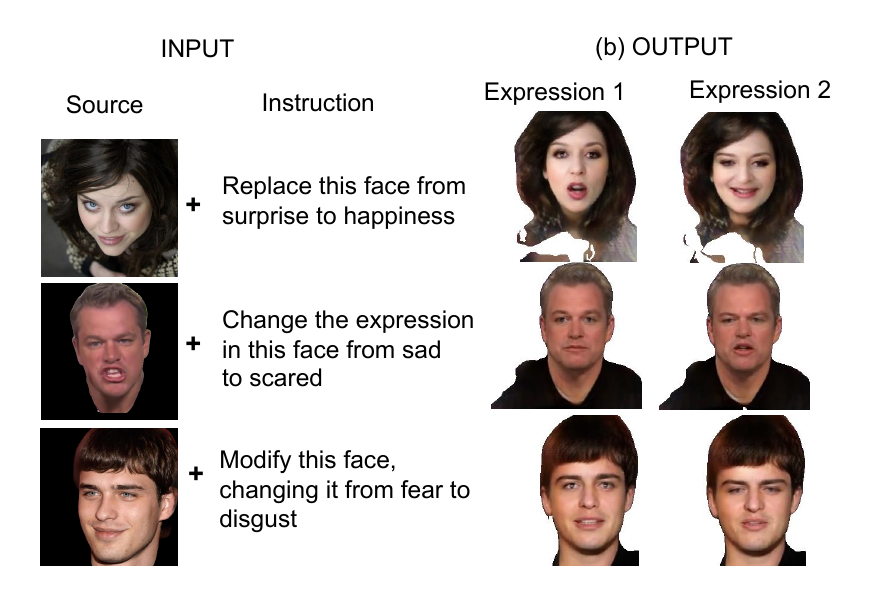}
    \caption{ }
    \label{fig:failure_cases}
\end{subfigure}
\belowcaptionskip = -15pt
\caption{Generated 3D faces (a) by changing the template sentences and (b) a few unsuccessful cases.}
\label{fig:t}
\end{figure}
\subsection{Discussion}
\subsubsection{Diversity of Facial Expression Sequences}
Our framework is capable of animating various facial expressions guided by instruction and facial images with pose variation, as illustrated in Fig \ref{fig:ck-long-sequence-with-neutral}. Moreover, it can generate facial expressions corresponding to both controlled and real-world environments, enabling the generation of a diverse range of expressions for creating a long video.
\subsubsection{Results on Instructions with Changes in Template Sentences}
Fig. \ref{fig:vocab} shows the outcomes obtained by changing the template sentences. The result suggests that our model can successfully generate facial expressions that closely resemble specific emotions described in the instruction.

\subsubsection{Failure Cases}
Although our model performs well in many cases, the model failed in certain instances as exemplified in Fig \ref{fig:failure_cases}. In the top row, significant pose variation in the input image leads to regions of the shoulders being missing from the resulting image due to constraints within the pre-trained face-rendering model. Likewise, images in which regions of the head and those containing hair were missing from the source image also led to the lack of information in the face rendering process because of this constraint, as shown in the bottom row.   
In the second row, the generated result is restricted by the size of the expression vocabulary, causing a \textit{scared} face to be transformed into an \textit{angry} face.
\section{Conclusion}
\label{sec:conclusion}
In this study, we present a novel framework for generating 3D facial expressions from an RGB image and animating facial transitions between two expressions as specified by entering text instructions.
First, FET was introduced to produce a sequence of facial expressions. The Instruction-Driven Facial Expression Decomposer was designed to learn from multimodal data and capture correlations between textual descriptions and facial expression features. We proposed an Instruction to Facial Expression Transition model to generate target facial expressions guided by a single instruction. Our framework integrates FET and a pre-trained face rendering model to generate facial appearances aligned with the expected expression sequence.
Furthermore, our framework can be expanded to include the generation of a neutral expression, thereby enabling the creation of diverse expressions in a video, thus closely simulating facial expression behaviors in real-world scenarios. Extensive experiments were conducted on the CK+ and CelebV-HQ datasets to demonstrate the effectiveness of our framework.    
Although our proposed approach demonstrates positive outcomes in terms of controlling the facial expression of an RGB image according to the provided instructions, it does have certain limitations. Similar to previous studies \cite{Khakhulin2022ROME, ma2023cvthead} in which facial parameters were used, the performance of our model relies on the precision of 3D face coefficients, especially using DECA \cite{DECA:Siggraph2021} within our configurations since it may encounter difficulty, often seen in challenging scenarios, in disentangling facial factors. Also, even though linear interpolation can maintain temporal consistency, it often compromises the realism of the synthesized videos. Although the present work uses either basic facial expressions or two designated expressions, the vocabulary number can be easily expanded by combining a Large Language Model (LLM) with our model. Given that text prompting is a powerful tool by which various emotional states can be expressed on a 3D avatar via our framework, we expect it to find various applications in the future.

\section*{Acknowledgment}
This work was supported by the Information Technology Research Center (ITRC) support program (IITP-2022-RS-2022-00156354) 
 and a Korean government grant (MSIT) (No.RS-2019-II190231) from the Institute of Information \& Communications Technology Planning \& Evaluation (IITP) as well as by the Basic Science Research Program through the National Research Foundation of Korea (NRF) funded by the Ministry of Education (2020R1A6A1A03038540).

\ifCLASSOPTIONcaptionsoff
  \newpage
\fi



%\bibliographystyle{IEEEtran}
%\begin{thebibliography}{1}


\bibliographystyle{IEEEtran}
\bibliography{bibtex/bib/references}
\end{document}